\documentclass[10pt,journal,compsoc]{IEEEtran}
\ifCLASSOPTIONcompsoc
  \usepackage[nocompress]{cite}
\else
  \usepackage{cite}
\fi
\ifCLASSINFOpdf
\else
\fi
\usepackage{hyperref} 

\usepackage{times}
\usepackage[pdftex]{graphicx}
\usepackage{amsmath}
\usepackage{amssymb}

\usepackage{soul}
\usepackage{subcaption}
\usepackage{caption}
\usepackage{tabularx}
\def\nbR{\ensuremath{\mathrm{I\! R}}}

\usepackage{multirow}
\usepackage{makecell}
\usepackage{comment}
\usepackage{pifont}
\usepackage{xcolor}
\usepackage{textcase}
\usepackage[accsupp]{axessibility}
\newcommand{\cmark}{\ding{51}}%
\newcommand{\xmark}{\ding{55}}%

\newcommand{\datasetname}{HowToVQA69M}
\newcommand{\webdataname}{WebVidVQA3M}
\newcommand{\smalldatasetname}{iVQA}

\newcommand{\vqat}{VQA-T}
\newcommand{\qat}{QA-T}

\newcommand{\etal}{\textit{et al.}}
\newcommand{\eg}{\textit{e.g.}}
\newcommand{\ie}{\textit{i.e.}}
\newcommand\titlelowercase[1]{\texorpdfstring{\lowercase{#1}}{#1}}

\begin{document}
%
\title{Learning to Answer Visual Questions \\ from Web Videos}
\author{Antoine Yang, Antoine Miech, Josef Sivic,
Ivan Laptev, Cordelia Schmid 
\IEEEcompsocitemizethanks{\IEEEcompsocthanksitem A. Yang, I. Laptev and C. Schmid are with Inria Paris and D\'{e}partement d'informatique de l'ENS, CNRS, PSL Research University, 75005 Paris, France.
}
\IEEEcompsocitemizethanks{\IEEEcompsocthanksitem A. Miech is with DeepMind, London, United Kingdom.
}
\IEEEcompsocitemizethanks{\IEEEcompsocthanksitem J. Sivic is with Czech Institute of Informatics, Robotics and Cybernetics,
Czech Technical University in Prague.
}
}

\IEEEtitleabstractindextext{%
\begin{abstract}
Recent methods for visual question answering rely on
large-scale annotated datasets. 
Manual annotation of questions and answers for videos, however, is tedious, expensive and prevents scalability. 
In this work, we propose to avoid manual annotation and generate a large-scale training dataset for video question answering making use of automatic cross-modal supervision. 
We leverage a question generation transformer trained on text data and use it to generate question-answer pairs from transcribed video narrations. 
Given narrated videos, we then automatically generate the HowToVQA69M dataset with 69M video-question-answer triplets. 
To handle the open vocabulary of diverse answers in this dataset, we propose a training procedure based on a contrastive loss between a video-question multi-modal transformer and an answer transformer.  
We introduce the zero-shot VideoQA task and the VideoQA feature probe evaluation setting and show excellent results, in particular for rare answers. 
Furthermore, our method achieves competitive results on MSRVTT-QA, ActivityNet-QA, MSVD-QA and How2QA datasets.
We also show that our VideoQA dataset generation approach generalizes to another source of web video and text data. We use our method to generate the \webdataname{} dataset from the WebVid dataset, i.e., videos with alt-text annotations, and show its benefits for training VideoQA models.
Finally, for a detailed evaluation we introduce \smalldatasetname{}, a new VideoQA dataset with reduced language bias and high-quality manual annotations. 
Code, datasets and trained models are available on our project webpage$^1$.
\end{abstract}

}

\maketitle

\footnotetext[1]{https://antoyang.github.io/just-ask.html}

\IEEEdisplaynontitleabstractindextext

%
\IEEEpeerreviewmaketitle

\IEEEraisesectionheading{\section{Introduction}\label{sec:intro}}
Answering questions about videos requires a detailed understanding of the visual content and its association with the natural language. 
Indeed, given the large diversity of questions, methods for Video Question Answering (VideoQA) should reason about  scenes, objects and human actions as well as their complex temporal interactions. 

Current approaches to VideoQA rely on deep fully-supervised models trained on manually annotated datasets with question and answer pairs~\cite{fan2019heterogeneous, huang2020location, jiang2020divide, jiang2020reasoning, le2020hierarchical, lei2021less, li2019beyond}.
Collecting and annotating VideoQA datasets, however, is cumbersome, time consuming, expensive and therefore not scalable. 
As a result, current VideoQA datasets are relatively small (see Figure~\ref{fig:datasetcomparison}). 
This limitation hinders the progress in the field as state-of-the-art VideoQA models often require a large amount of training data.

\begin{figure}[t]
\centering
\includegraphics[width=1.\linewidth]{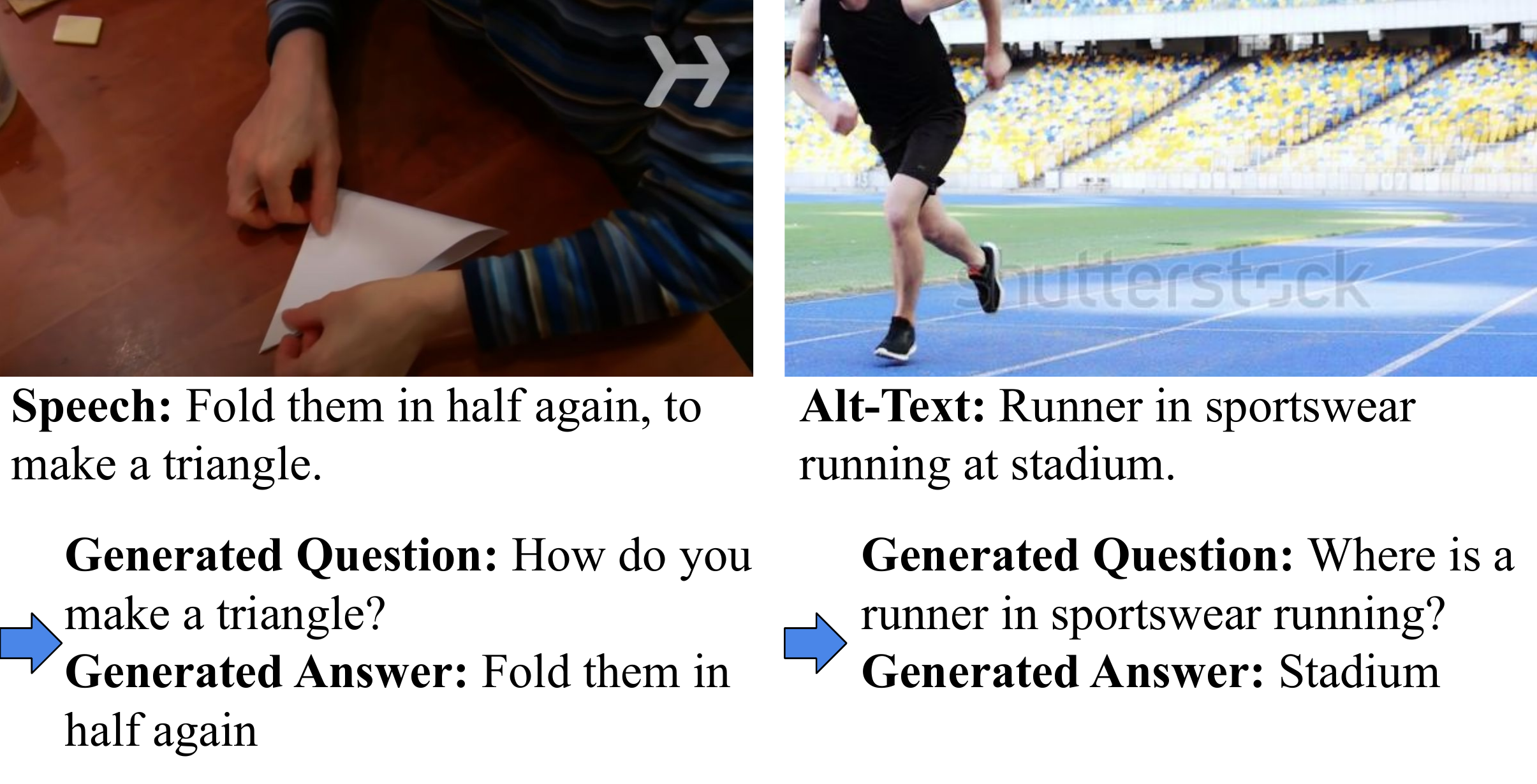}
\vspace{-0.5cm}
\caption{\small Given videos with transcribed narration (left) or videos with ``alt-text" annotations (right), we leverage language models and cross-modal supervision to obtain large-scale VideoQA data. Top: Example frame with the corresponding text annotation. Bottom: automatically generated question and answer pair. 
}
\label{fig:teasersqa}
\vspace{-0.5cm}
\end{figure}

\begin{figure*}[t]
\centering
\includegraphics[width=\linewidth]{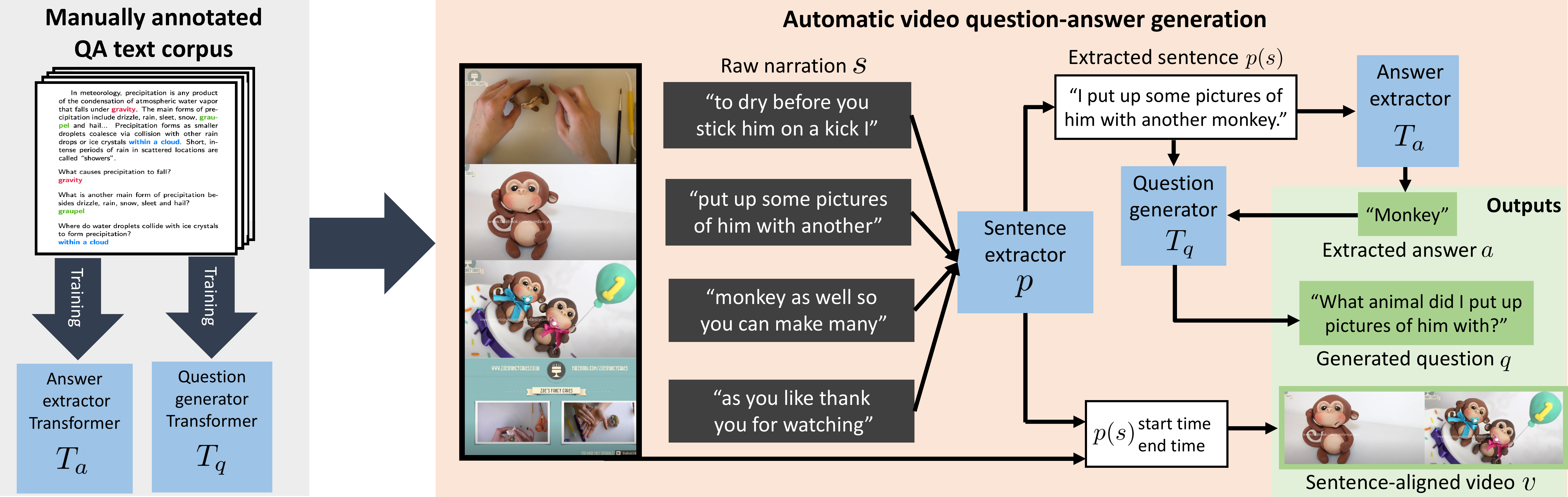}
\vspace{-0.5cm}
\caption{\small {\bf Our automatic approach for large-scale generation of video-question-answer triplets from narrated (subtitled) videos.} First, at the language-only training phase (left), the transformer-based answer extractor $T_a$ and question generator $T_q$ are trained~\cite{raffel2020exploring} on a manually annotated text-only question-answer corpus. Then video-question-answer triplets are automatically generated from narrated videos (right). Individual sentences are extracted from the ASR-transcribed narration using a punctuator $p$. Each extracted sentence is analyzed with an answer extractor $T_a$ and a question generator $T_q$ to produce answer $a$ and question $q$. The timestamps of the narration are used to obtain a video clip $v$ temporarily aligned to the extracted sentence to form the output video-question-answer triplet~$(v, q, a)$.}
\vspace{-0.3cm}
\label{fig:data_generation}
\end{figure*}

In this work, we address the scale issue with a new approach for automatically generating  VideoQA datasets as illustrated in Figure~\ref{fig:teasersqa}. 
The idea is to leverage cross-modal supervision together with text-only tools for question generation and to automatically annotate VideoQA data from a \emph{large amount of videos with readily-available text annotations} in the form of transcribed narrations or ``alt-text" annotations available with the video on the Internet.
Inspired by the recent progress in language generation using transformer-based language models~\cite{brown2020language}, we leverage transformers trained on a question-answering text corpus to generate a diverse set of non-scripted questions and corresponding open-vocabulary answers from text. 
By applying these transformers to speech transcripts of narrated videos from the large-scale HowTo100M dataset~\cite{miech19howto100m} we create \datasetname{}, an open-ended VideoQA dataset with 69 million video-question-answer triplets and a diverse set of more than 16M unique answers (see Figure~\ref{fig:data_generation}). 
We also extend our approach to web videos with readily-available alt-text descriptions and generate the \webdataname{} dataset from the WebVid2M dataset~\cite{bain2021frozen}.
As shown in Figure~\ref{fig:datasetcomparison}, our \datasetname{} and \webdataname{} datasets are orders of magnitude larger compared to prior VideoQA datasets.

Given the limited diversity of existing datasets, current methods typically reduce VideoQA to a classification problem, where frequent answers are assigned to unique classes. Typically, up to 5K unique possible answers are considered. 
Such an approach, however, does not scale to the open vocabulary of 16M different answers in \datasetname{}.
To address this problem and to enable video question answering with highly diverse questions and answers, we introduce a training procedure based on contrastive learning  between a video-question multi-modal  transformer and an answer transformer that can handle free-form answers. This bypasses the need to define a discrete set of answer classes.

The goal of our work is to advance truly open-ended and generic solutions to VideoQA. To evaluate generalization,  we propose a new zero-shot VideoQA task 
where we prohibit any manual supervision of visual data during training, and a new VideoQA feature probe evaluation setting where only the final projection layers of the network are finetuned on the target dataset. 
Our VideoQA model, trained on~\datasetname{}, demonstrates excellent zero-shot results on multiple existing datasets, especially for rare answers. 
Additionally, we find that our VideoQA model exhibits strong performance in the VideoQA feature probe evaluation setting.
Moreover, when finetuned on target datasets, our model achieves competitive results on
MSRVTT-QA~\cite{xu2017video}, ActivityNet-QA~\cite{yu2019activitynet}, MSVD-QA~\cite{xu2017video} and How2QA~\cite{li2020hero}.
We further show the generalizability of our approach by showing the benefits of \webdataname{} for training VideoQA models.

Initial experiments have shown that existing benchmarks for open-ended VideoQA~\cite{xu2017video, yu2019activitynet} contain a language bias~\cite{goyal2017making}, i.e., their questions can often be answered without looking at the video. 
To better evaluate the impact of visual information in VideoQA, we introduce a new open-ended VideoQA dataset (iVQA) with manually collected questions and answers, where we exclude questions that could be answered without watching the video. 
Moreover, to account for multiple possible answers, iVQA contains five independently collected answers for each question.

In summary, our work makes the following three contributions: 
\vspace{-.6cm}
\begin{itemize}
    \item[\textit{(i)}]
We introduce an approach to automatically generate a large-scale VideoQA dataset, \datasetname. Relying on cross-modal supervision, we use transformers trained on an existing text-only question-answering corpus and generate video-question-answer triplets from videos and transcribed narrations. 
We also apply our method to video alt-text pairs and generate the \webdataname{} dataset.
    \item[\textit{(ii)}]
We train a VideoQA model on the automatically generated data via contrastive learning between a multi-modal video-question transformer and an answer transformer. We show the efficiency of our model for the new zero-shot VideoQA task and the new VideoQA feature probe task. Our model achieves competitive results in four existing VideoQA benchmarks. 
    \item[\textit{(iii)}]
Finally, we  introduce a new manually annotated open-ended VideoQA benchmark \smalldatasetname{} that excludes non-visual questions and contains multiple possible answers for each question.
\vspace{-.2cm}
\end{itemize}

 \noindent
 Code, datasets and trained models are available at {\small{\url{https://antoyang.github.io/just-ask.html}}}.

\section{Related Work}\label{sec:background}
\noindent \textbf{Visual Question Answering (VQA).}
VQA is typically tackled by classifying the  image-question (or video-question) representation into a fixed vocabulary of answers.
Various approaches to combine spatial image representations and sequential question representations have been proposed~\cite{anderson2018bottomup, ben2017mutan, fukui2016multimodal, lu2016hierarchical, xiong2016dynamic, xu2016ask, yang2016stacked}. More specifically to the video domain (VideoQA), spatio-temporal video representations in terms of motion and appearance have been used in~\cite{dang2021object, fan2019heterogeneous, gao2018motion, huang2020location, jang2017tgif, jiang2020divide, jiang2020reasoning, le2020hierarchical, le2020neural, lei2021less, li2019beyond, park2021bridge, seo2021attend, xu2017video, xue2018better, yu2021learning, zha2019spatiotemporal, zhuang2020multichannel}. 

\begin{figure}[t]
\centering
\includegraphics[width=\linewidth]{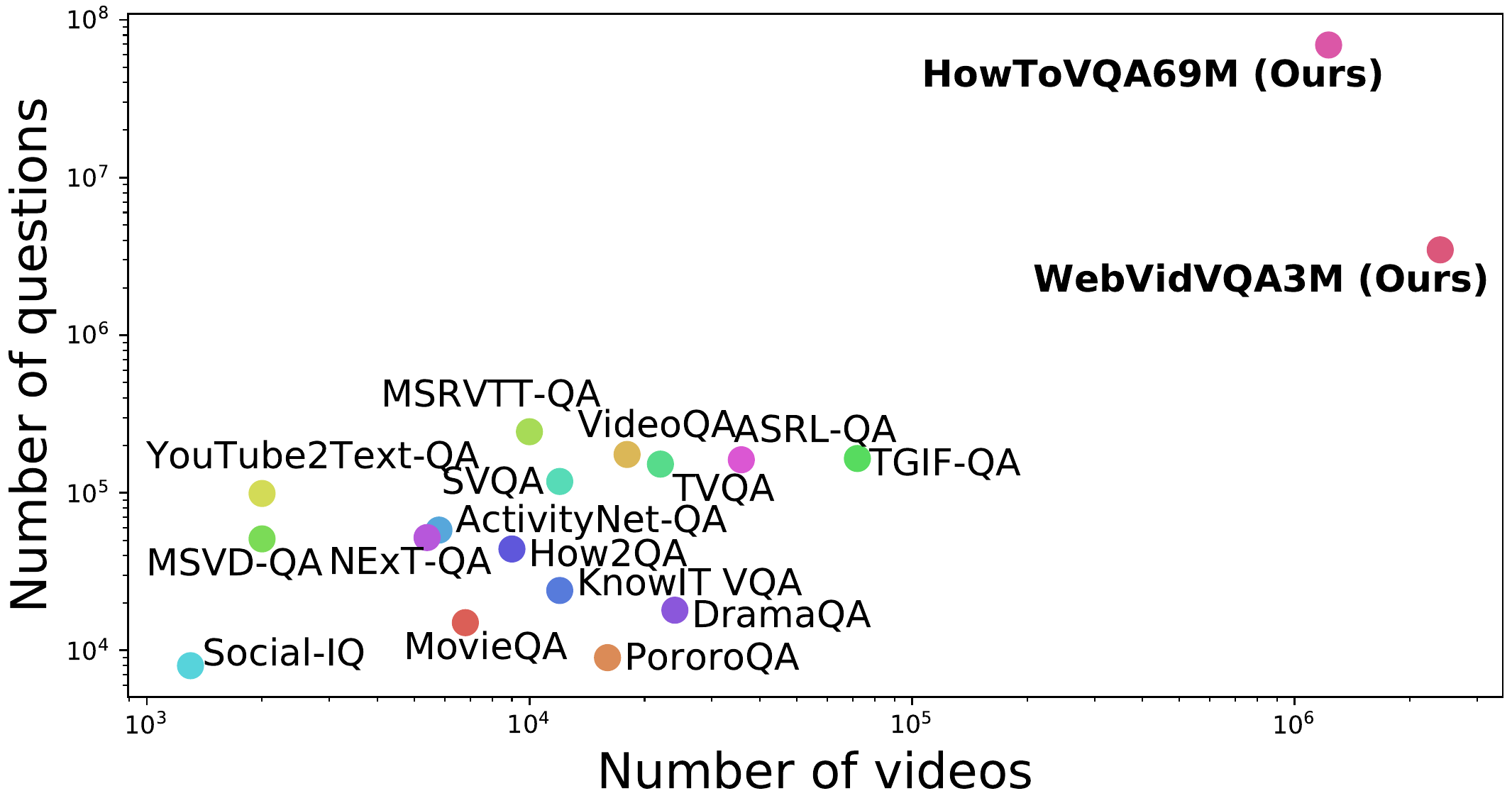}
\vspace{-0.5cm}
\caption{\small Comparison of our large-scale \datasetname{} and \webdataname{} datasets with existing VideoQA datasets.}
\vspace{-0.5cm}
\label{fig:datasetcomparison}
\end{figure}

Methods above are limited to pre-defined vocabularies of answers and are difficult to apply outside of specific datasets.
To address this problem, Hu \etal~\cite{hu2018learning} propose a joint embedding where image-question representations can be matched with free-form answers.
Our VideoQA model follows this idea, but instead of relying on manually annotated datasets of limited scale, we train it on a large-scale VideoQA dataset that we automatically generate.
In contrast to some previous works using additional video features such as subtitles~\cite{chadha2020iperceive, kim2020dense, kim2020modality, kim2021self, lei2018tvqa, lei2019tvqa+, li2020hero, lin2021vx2text, tapaswi16movieqa, winterbottom2020modality,yang2020bert}, our video representation is exclusively based on visual information, as we focus on the detailed visual understanding of videos.

To evaluate the generalization of VQA models, Teney and Hengel~\cite{teney2016zero} define zero-shot VQA by answering previously unseen questions, which is a related but less challenging task compared to the zero-shot VQA task we propose in Section~\ref{sec:zeroshot}. Vatashsky and Ullman~\cite{vatashsky2020vqa} address VQA using COCO image annotations \cite{lin14coco}, while our zero-shot model is trained with no manual annotations. Our proposed zero-shot VQA task is analogous to zero-shot video retrieval~\cite{miech20endtoend} or zero-shot action recognition~\cite{radford2021learning}. 
We further propose a VQA feature probe evaluation setting where only the final heads of the network are finetuned on the downstream dataset while all other pretrained weights are kept frozen. 
This setting is analogous to the linear probe evaluation setting commonly used in self-supervised image recognition~\cite{caron2020unsupervised, caron2021emerging, chen2021empirical} or self-supervised action recognition~\cite{radford2021learning} but with multiple layers in the head rather than just a single (linear) layer.  

Visual question generation (VQG) has been introduced in \cite{mostafazadeh2016generating}. The methods in~\cite{li2018visual} and \cite{shah2019cycle} propose to jointly learn VQG and VQA to improve the image VQA task. However, these works do not generate questions to obtain additional training data, but use visual data annotation for VQG as an additional loss. 

\vspace*{1mm}
\noindent \textbf{VideoQA datasets.}
Manually collecting and annotating video-question-answer triplets is cumbersome, costly and difficult to scale.
As a result, current VideoQA datasets~\cite{castro2020lifeqa, choi2020dramaqa, colas2019tutorialvqa,fan2019egovqa,garcia2020knowit,jang2017tgif,kim2017deepstory, lei2018tvqa,li2020hero,mun2017marioqa, sadhu2021video, song2018explore, tapaswi16movieqa, xiao2021next, xu2017video,ye2017video,yu2019activitynet, zadeh2019social, zeng2017leveraging} are limited in size, as the largest, TGIF-QA~\cite{jang2017tgif}, contains only 72K annotated clips (see Figure~\ref{fig:datasetcomparison} for more details).
To address this issue, several works have explored leveraging manually annotated video descriptions~\cite{jang2017tgif, wang2020long, xu2017video, zeng2017leveraging, zhao2020open, zhao2017video, zhao2018open} for automatic generation of VideoQA datasets, using rule-based~\cite{heilman2010good, ren2015exploring} approaches. Similarly, in the image domain, Banerjee \etal~\cite{banerjee2021weaqa} has recently proposed to use annotated image captions from COCO~\cite{chen2015microsoft} to generate question-answer pairs using a template-based approach~\cite{ren2015exploring}.

Instead, we propose to use video annotations in the form of transcribed narrations or alt-text descriptions that are available at large-scale with no manual supervision. Moreover, rule-based generation requires the manual creation of rules by experts which is expensive, and has also been recently outperformed by neural question generation~\cite{du2017learning, yao2018teaching, zhou2017neural} as used in our approach.

\vspace*{1mm}
\noindent \textbf{Large-scale pretraining for vision and language.}
Several recent methods~\cite{alberti2019fusion, chen2019uniter, desai2020virtex, huang2020pixel, huang2021seeing, li2019unicodervl, li2019visualbert, li2020oscar, lu2019vilbert, lu202012, su2019vl, tan2019lxmert, tubedetr, zhou2020unified} pretrain multi-modal vision-language representations, such as transformers, using datasets with image captions, e.g., COCO~\cite{chen2015microsoft}, Conceptual Captions~\cite{sharma2018conceptual} and Visual Genome~\cite{visualgenome}.
These methods are often optimized using generic objectives such as masked language losses and losses for text-image matching and image caption generation.
In our work, we pretrain models using large amounts of narrated videos. In contrast to task-agnostic pretraining in the previous work, we show the benefits of task-specific pretraining for our target VideoQA task.

\vspace*{1mm}
\noindent \textbf{Learning from Web videos.} 
In this work, we exploit noisy correlations between videos and readily-available text annotations in unlabeled web videos from the recent HowTo100M~\cite{miech19howto100m} and WebVid2M~\cite{bain2021frozen} datasets.
Methods using such readily-available data have shown significant improvements on several tasks including video retrieval~\cite{bain2021frozen,gabeur2020multi}, action localization~\cite{miech19howto100m}, action recognition~\cite{miech20endtoend} and video captioning~\cite{luo2020univilm,sun2019contrastive,sun2019videobert,zhu2020actbert}, sometimes outperforming fully-supervised baselines.
Others have used videos with readily available text annotations for the VideoQA task.
In detail, Amrani \etal~\cite{amrani2020noise} propose a text-video pretraining approach and finetune their model for VideoQA.
Li \etal~\cite{li2020hero} propose HERO, a pretraining approach restricted to multiple-choice VideoQA, for which questions and answers are treated as a single text stream.
Seo \etal~\cite{seo2020look} propose a pretraining approach based on next utterance prediction and finetune their model for VideoQA. 
Zellers \etal~\cite{seo2020look} propose a pretraining approach based on a mix of frame-level and video-level objectives and finetune for VideoQA.
Differently to these methods with task-agnostic pretraining, we propose a pretraining approach specifically dedicated for VideoQA using automatically generated question and answer pairs from readily available text annotations, and show in Section~\ref{sec:experiments} the benefits of our approach.

A preliminary version of this article has appeared in~\cite{yang2021just}.

\section{Large-scale generation of VideoQA data}\label{sec:generation}
\begin{figure*}[t]
\begin{center}
\includegraphics[width=1.\linewidth]{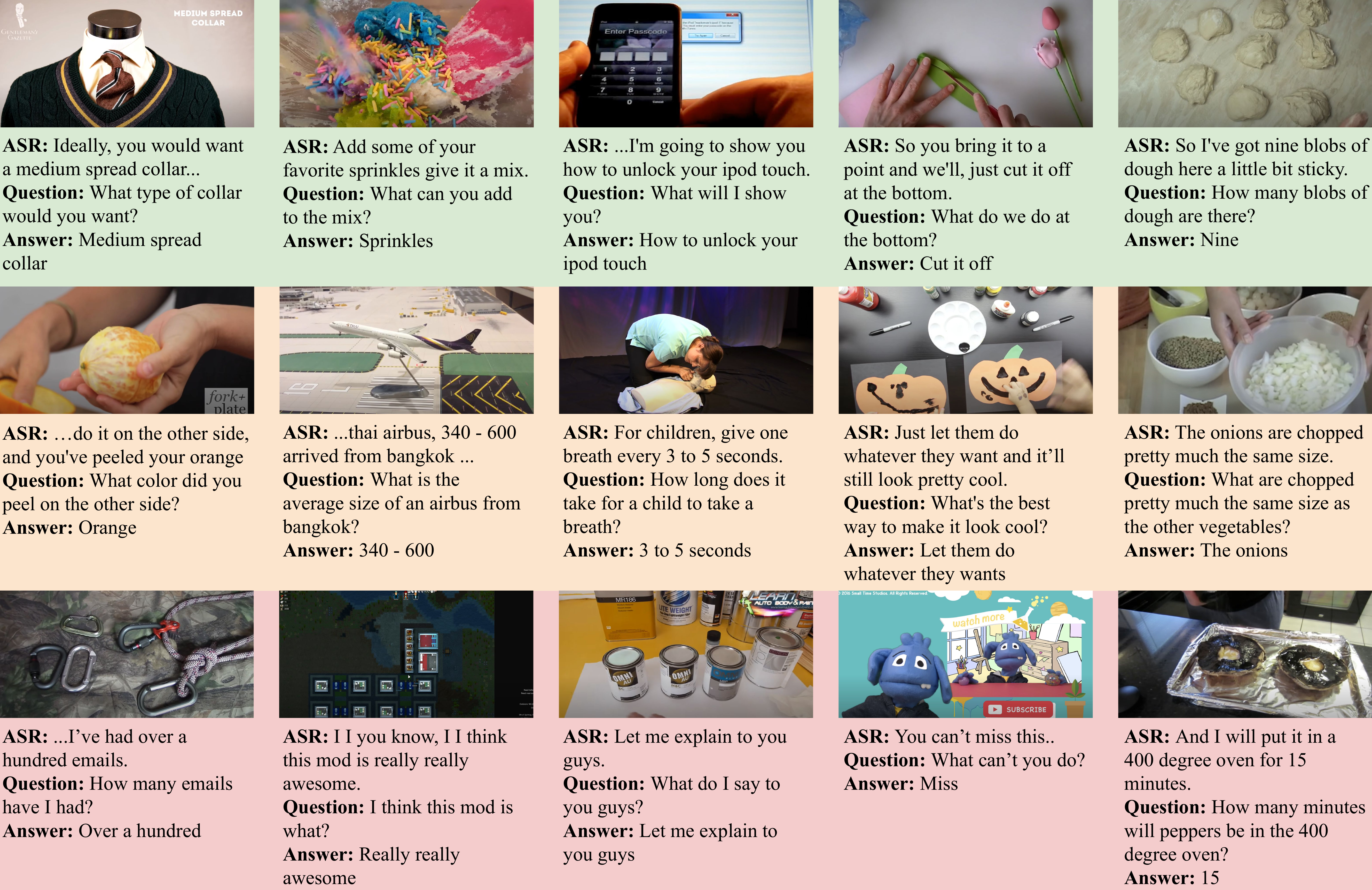}
\end{center}
\vspace{-0.4cm}
\caption{\small Examples of video-question-answer triplets generated from narrated videos in our \datasetname{} dataset. {\color{green}The green color} (first row) indicates relevant examples, {\color{orange}the orange color} (second row) indicates a failure of the question-answer generation, and {\color{red}the red color} (third row) indicates that the generated question-answer is unrelated to the visual content.}
\vspace{-0.2cm}
\label{fig:\datasetname{}}
\end{figure*}

\begin{figure}[t]
\centering
\begin{subfigure}{0.49\linewidth}
\includegraphics[width=\linewidth]{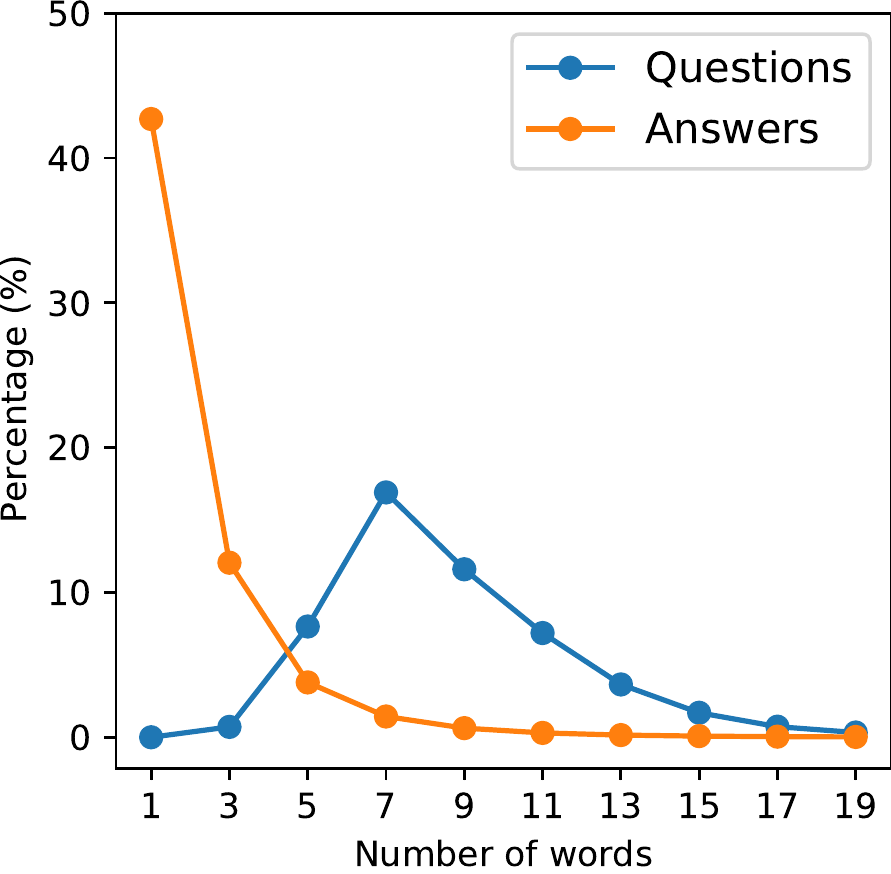}
\caption{Question and answer length}
\end{subfigure}%
\hfill
\begin{subfigure}{0.49\linewidth}
\includegraphics[width=\linewidth]{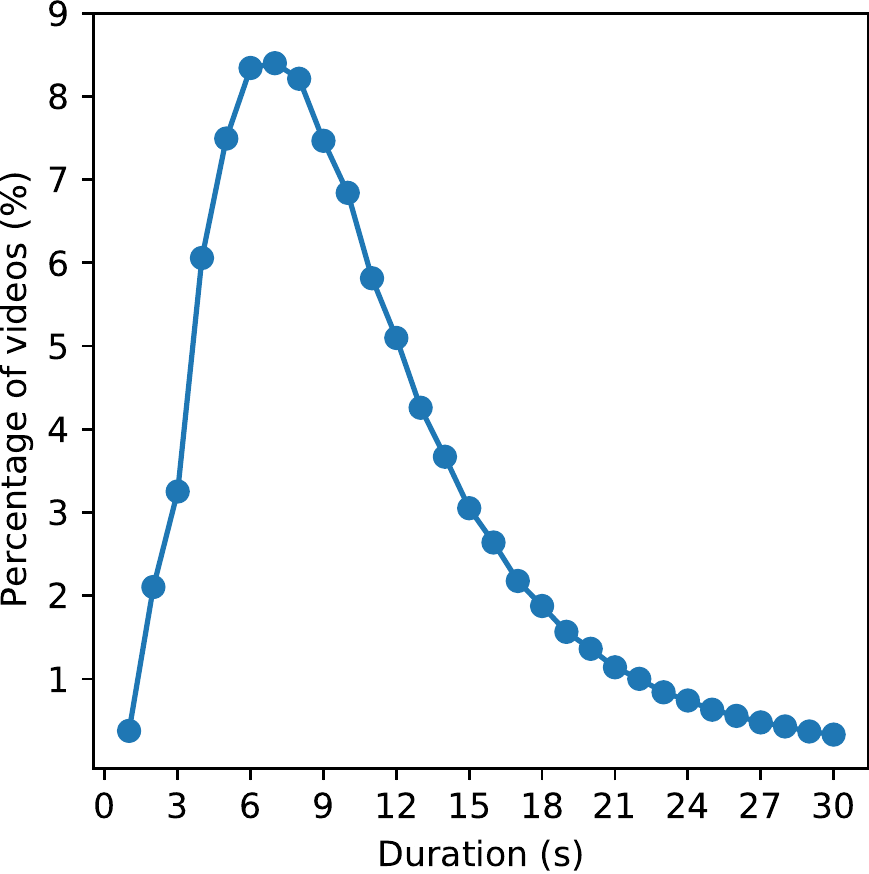}
\caption{Clip duration}
\end{subfigure}
\vspace{-0.2cm}
\caption{{\bf Statistics of the \datasetname{} dataset.} (a)~Distribution of length of questions and answers. (b)~Distribution of video clip duration in seconds.}
\label{fig:sqa_length}
\vspace{-0.3cm}
\end{figure}

\begin{figure}[t]
\centering
\begin{subfigure}{\linewidth}
\includegraphics[width=\linewidth]{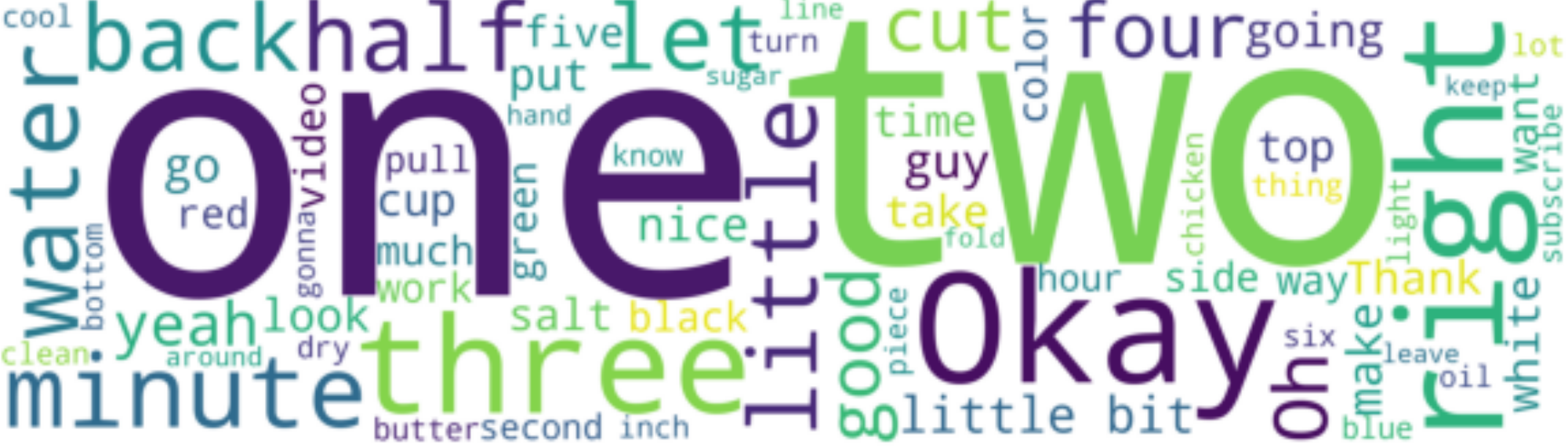}
\caption{Answers}
\end{subfigure}
\begin{subfigure}{\linewidth}
\includegraphics[width=\linewidth]{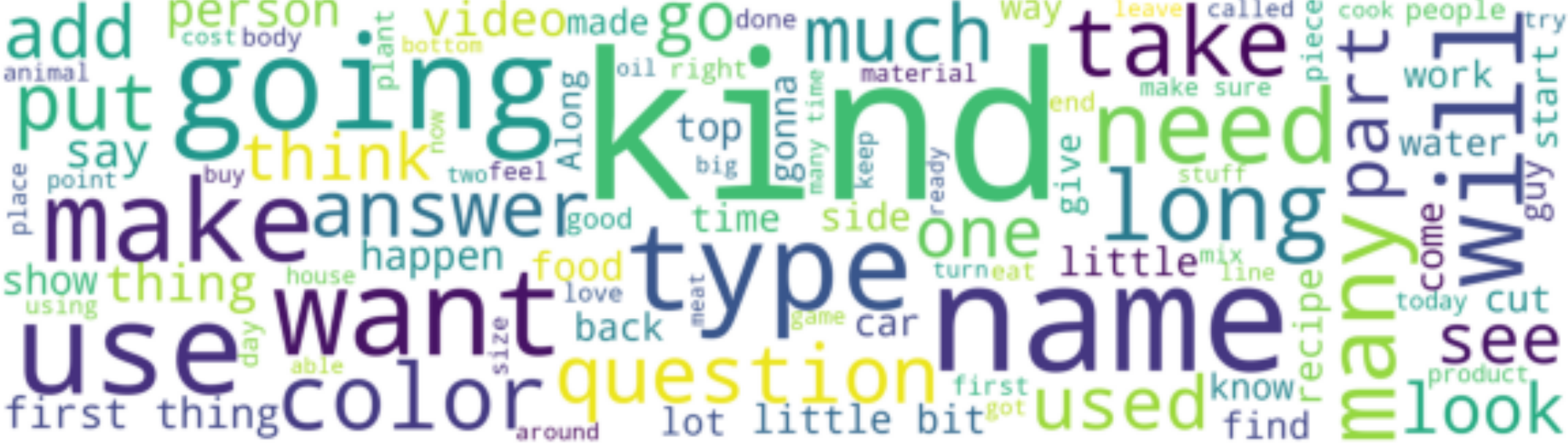}
\caption{Questions}
\end{subfigure}
\vspace{-0.2cm}
\caption{Word clouds extracted from the \datasetname{} dataset showing its diverse vocabulary and the words characteristic to speech such as \textit{okay}, \textit{right}, or \textit{oh}.}
\label{fig:sqa_words}
\vspace{-0.5cm}
\end{figure}

This section presents our approach to generate large-scale VideoQA datasets from videos with readily available text annotations. 
We illustrate the proposed approach on instructional videos with text annotations in the form of transcribed narrations, which in many cases describe the content of the videos. 
Section~\ref{sec:qgen} presents our proposed generation procedures. Section~\ref{sec:\datasetname{}}, then, describes the resulting \datasetname{} dataset.
Our approach can be easily adapted to other type of content, for example, shorter web-videos with with readily text annotations in the form of alt-text, as will be shown in the result section (Section~\ref{sec:results}).  

\subsection{Generating video-question-answer triplets}\label{sec:qgen}
We tackle the task of generating video-question-answer triplets from a large-scale instructional video dataset with transcribed spoken narration~\cite{miech19howto100m}. 
This is a challenging task because of transcription errors and lack of punctuation. We also wish to obtain highly diverse data. To address these issues, we propose to leverage powerful language models trained on text data. Our approach is illustrated in Figure~\ref{fig:data_generation} and details are given next. 

We first present details about the generation procedure. Let $s$ be the transcribed speech data obtained with automatic speech recognition (ASR). First, we use a recurrent neural network $p$, to infer punctuation in the transcribed speech data.
We denote the punctuated transcript as $p(s)$. 
We extract video clips $v$ temporally aligned with the inferred sentences $p(s)$ using the ASR timestamps. 
We found that the generation works significantly better when applied to sentences rather than the original sentence fragments from the HowTo100M dataset, see Table~\ref{table:manual}.
Second, for each sentence, we apply a transformer $T_a$, to extract a set of potential answers:  $a = T_a(p(s))$.
Third, we use another transformer $T_q$ to generate a question given each transcript sentence and each extracted answer such that: $q = T_q(a, p(s))$. 
The output is a set of video-question-answer triplets $(v, q, a)$.

We now explain details of the language models and their training procedure. For ASR, we follow \cite{miech19howto100m} and use the readily-available ASR data provided by YouTube. For punctuation $p$, we use the BRNN model from~\cite{tilk2016} and the weights available at~\cite{punct} trained on IWSLT2011~\cite{federico2012iwslt}. 
For $T_a$ and $T_q$, we use the transformer-based T5-small and T5-base models~\cite{raffel2020exploring}, respectively. 
We follow~\cite{alberti2019synthetic, chan2019recurrent, lopez2020transformer} and use the weights available at~\cite{qgen} trained for answer span extraction and answer-aware question generation, respectively, on SQuADv1~\cite{rajpurkar2016squad}.
SQuADv1 is a text-only question-answering dataset consisting of questions for which the answer is a segment of text extracted from a paragraph.

\subsection{\datasetname{}: a large-scale VideoQA dataset}\label{sec:\datasetname{}}
We have applied the previously described procedure to all 1.2M original videos from the HowTo100M dataset~\cite{miech19howto100m}.
The result is \datasetname{}, a dataset of 69,270,581 video clip, question and answer triplets $(v, q, a)$.
\datasetname{} is two orders of magnitude larger than any of the currently available VideoQA datasets (see Figure~\ref{fig:datasetcomparison}).
On average, each original video results in 43 video clips, where each clip is associated to 1.2 question-answer pairs. 
\datasetname{} is highly diverse and contains over 16M unique answers, where over 2M unique answers appear more than once and over 300K unique answers appear more than ten times. 
Examples of $(v, q, a)$ triplets from the \datasetname{} dataset are illustrated in Figure \ref{fig:\datasetname{}}, showing the diversity and the noise in the automatically obtained annotations in \datasetname{}.

\vspace*{1mm}
\noindent \textbf{Statistical analysis of \datasetname{}.}\label{sec:sqaanalysis}
Figure \ref{fig:sqa_length} shows the statistics of the \datasetname{} dataset in terms of the question length, answer length and video clip duration. 
Questions and answers contain 8.7 and 2.4 words on average respectively. 
Overall, \datasetname{} contains longer answers than downstream VideoQA datasets like MSRVTT-QA, MSVD-QA or ActivityNet-QA, for which answers are on average shorter than 2 words.
Each clip lasts 12.1 seconds on average.
The distribution of clip duration has a peak at around seven seconds with a long tail of longer clips.  These statistics demonstrate the diversity of our \datasetname{} dataset, in terms of videos, questions and answers.

Word clouds\footnote{Word clouds were generated using \url{github.com/amueller/word_cloud}.} for questions and answers in \datasetname{} are shown in Figure~\ref{fig:sqa_words} and illustrate the diverse vocabulary in \datasetname{} as well as the presence of speech-related words such as as \textit{okay}, \textit{right}, \textit{oh}. 

\begin{table}[t]
\begin{center}
\setlength\tabcolsep{2pt}
\resizebox{\linewidth}{!}{	
\begin{tabular}{ll|ccc}
Punctuation & Generation method & \makecell{ 
Correct \\ Samples} & \makecell{ QA Generation \\ Failure} & \makecell{ QA unrelated \\ to video} \\ 
\hline
\cmark & Heilman \etal~\cite{heilman2010good} & 17 & 54 & 29 \\
\xmark & Ours & 23 & 49 & 28 \\
\cmark & Ours & \textbf{30} & 31 & 39 \\
\end{tabular}
}
\end{center}
\vspace{-0.4cm}
\caption{\small Manual evaluation of our generation method (with and without punctuation) on a random sample of 100 examples compared with a rule-based question-answer generation of~\cite{heilman2010good}. Numbers are obtained with majority voting between 5 annotators.}
\vspace{-0.0cm}
\label{table:manual}
\end{table} 

\begin{table}[t]
\begin{center}
\setlength\tabcolsep{2pt}
\resizebox{0.85\linewidth}{!}{%
\begin{tabular}{ll|ccc}
\makecell{ Question \\ Type} & Total &
\makecell{ Correct \\ Samples (\%)} & \makecell{ QA Generation \\ Failure (\%)} & \makecell{ QA unrelated \\ to video (\%)} \\ 
\hline
Attribute & 25 & 28 & 32 & 40 \\
Object & 17 & 41 & 24 & 35 \\
Action & 16 & \textbf{69} & 19 & 13 \\
Counting & 13 & 23 & 15 & \textbf{62} \\
Place & 7 & 0 & \textbf{86} & 14 \\
People & 7 & 0 & 43 & 57 \\
Other & 15 & 13 & 27 & 60 \\
\end{tabular}%
}
\end{center}
\vspace{-0.4cm}
\caption{\small Manual evaluation of our generation method on 100 randomly chosen generated examples split by question type. Results are obtained by majority voting among 5 annotators.}
\vspace{-0.5cm}
\label{table:manualsplit}
\end{table} 

\vspace*{1mm}
\noindent \textbf{Manual evaluation of \datasetname{}.}\label{sec:eval}
As shown in Figure \ref{fig:\datasetname{}}, \datasetname{} annotations are noisy, which can be attributed to: (i) errors in speech transcription, (ii) speech not describing the video content, or (iii) errors in question-answer generation. 
We manually evaluate the quality of 100 randomly sampled $(v, q, a)$ triplets in \datasetname{} by collecting 5 different annotations for each triplet to reduce variance and report results in Table~\ref{table:manual}. 
Among 100 triplets generated by our method we find 30 to be  correctly generated and matching well to the video content, 31 are incorrectly generated and 39 are correctly generated but unrelated to the video content. 
To demonstrate the influence of the different components of our automatic question-answer generation procedure, we compare our results with (i) a variant of our approach that does not split transcribed narrations into sentences using a punctuator, and (ii) a rule-based approach~\cite{heilman2010good} for question-answer generation.
Table~\ref{table:manual} confirms the importance of punctuation and demonstrates the superior performance of our generation method compared to~\cite{heilman2010good}. Further comparison with~\cite{heilman2010good} is given in Section~\ref{sec:java}. 
In terms of inter-rater agreement, for the 300 generated video-question-answer triplets (100 for each generation method), 94 were in an agreement of all 5 annotators, 198 in an agreement of at least 4 annotators, and 299 in an agreement of at least 3 annotators. 
This high agreement of annotators demonstrates the reliability of the results in Table \ref{table:manual}.

We further manually classify the 100 video-question-answer triplets obtained with our method by the question type (``Attribute", ``Object", ``Action", ``Counting", ``Place", ``People", or ``Other"), evaluate the quality of generated triplets for different question types and report results in Table~\ref{table:manualsplit}.
Out of the 6 most common categories, we observe that questions related to ``Action" lead to the best annotations, ``Counting" questions lead to the highest number of QAs unrelated to the video content, and questions related to ``Place" lead to the highest number of QA generation errors. 
Qualitatively, we found that actions are often depicted in the video, while counted quantities (\eg~time, weight, length) mentioned in the speech are hard to guess from the video only. 
We describe next how we use \datasetname{} to train our VideoQA model.

\begin{figure}[t]
\centering
\includegraphics[width=1.\linewidth]{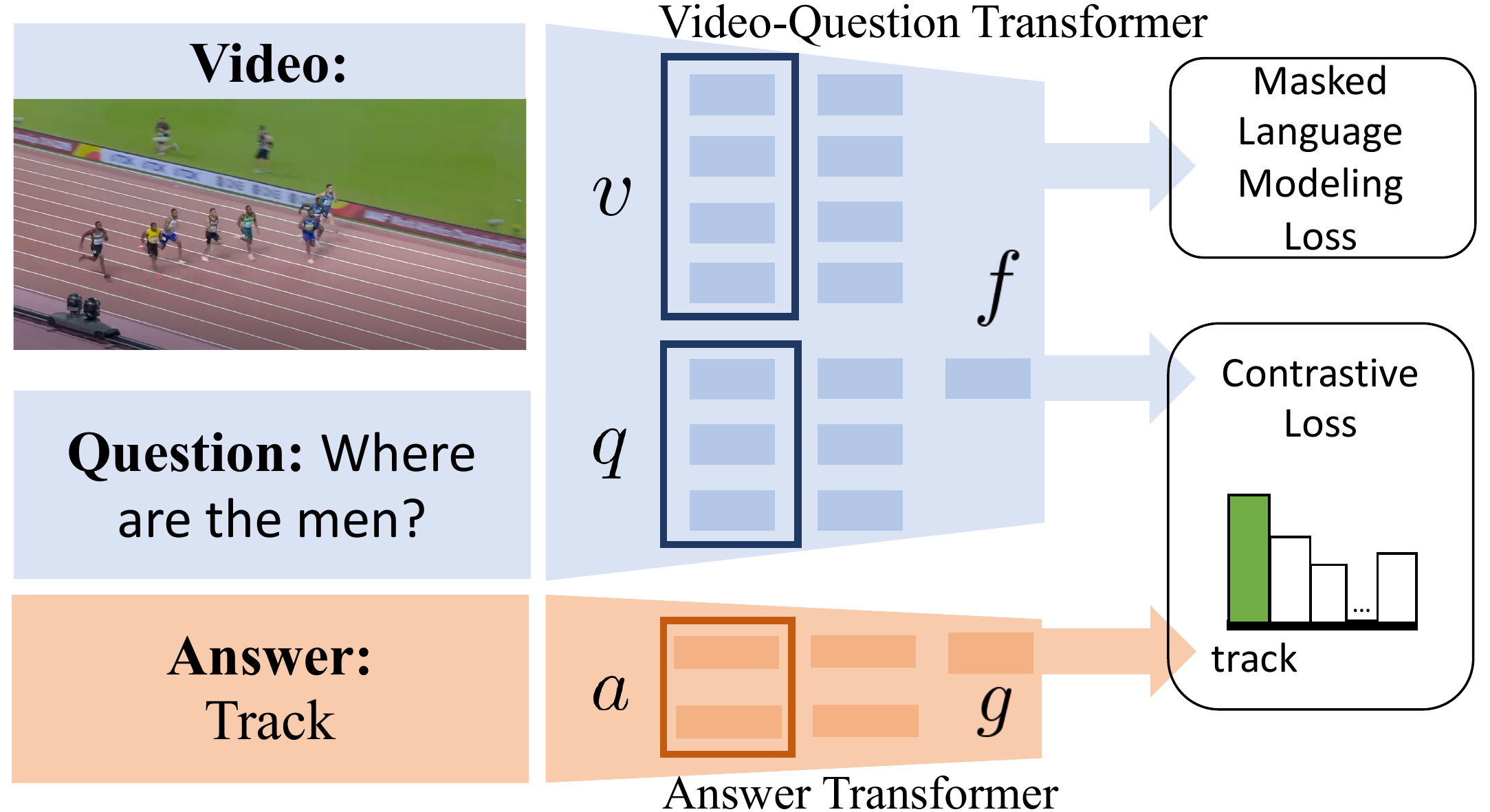}
\vspace{-0.5cm}
\caption{{\bf Overview of our VideoQA training architecture.} Our model is composed of a video-question module $f$ based on a multi-modal transformer (top) and an answer module $g$ based on DistilBERT \cite{sanh2019distilbert} encoder (bottom). For pretraining, we use a contrastive loss and a masked language modeling loss (right).}
\vspace{-0.5cm}
\label{fig:videoqamodel}
\end{figure}

\begin{figure*}[t]
\centering
\begin{subfigure}{0.24\linewidth}
\includegraphics[width=\linewidth]{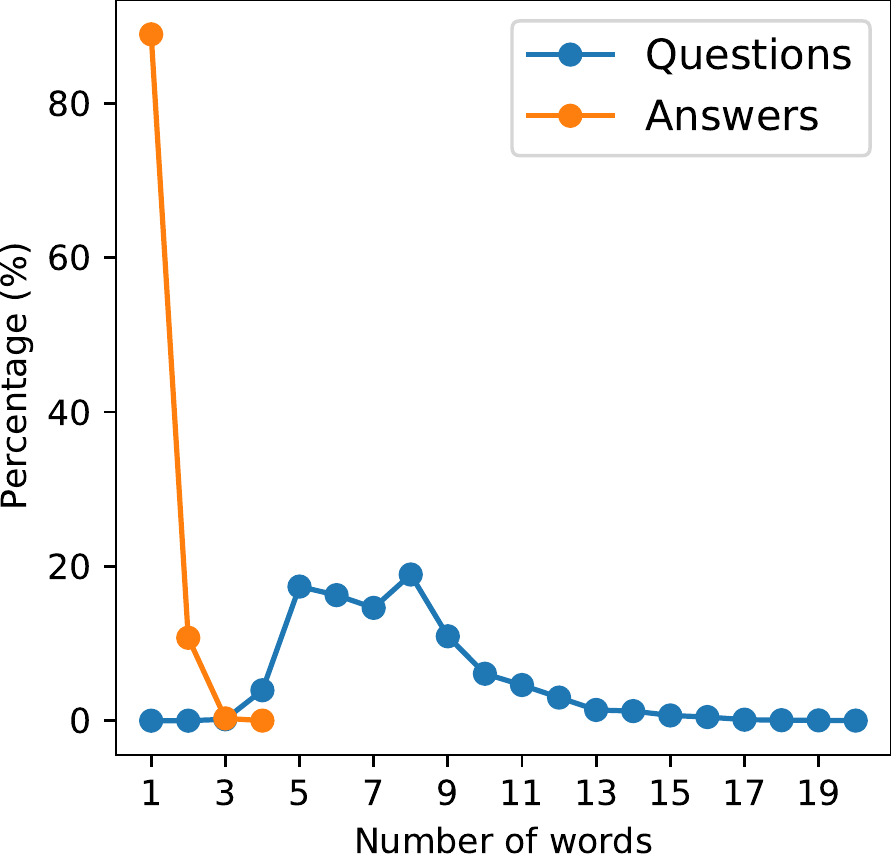}
\caption{Question and answer length}
\end{subfigure}%
\hfill
\begin{subfigure}{0.24\linewidth}
\includegraphics[width=\linewidth]{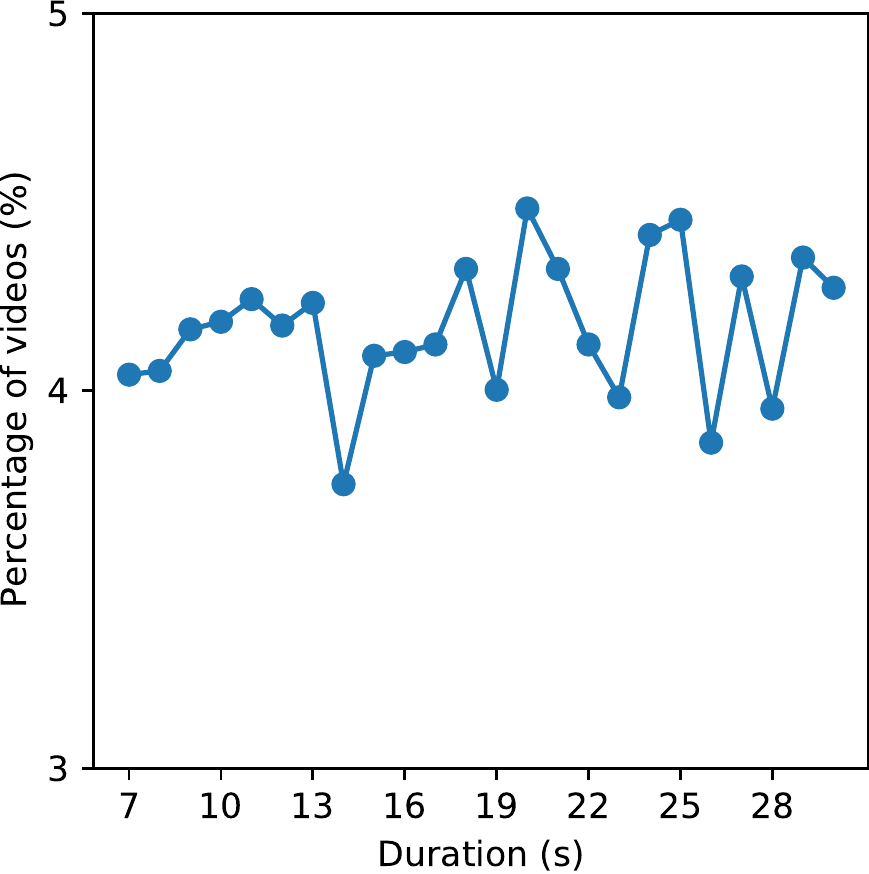}
\vspace{-0.5cm}
\caption{Clip duration}
\end{subfigure}%
\hfill
\begin{subfigure}{0.24\linewidth}
\includegraphics[width=\linewidth]{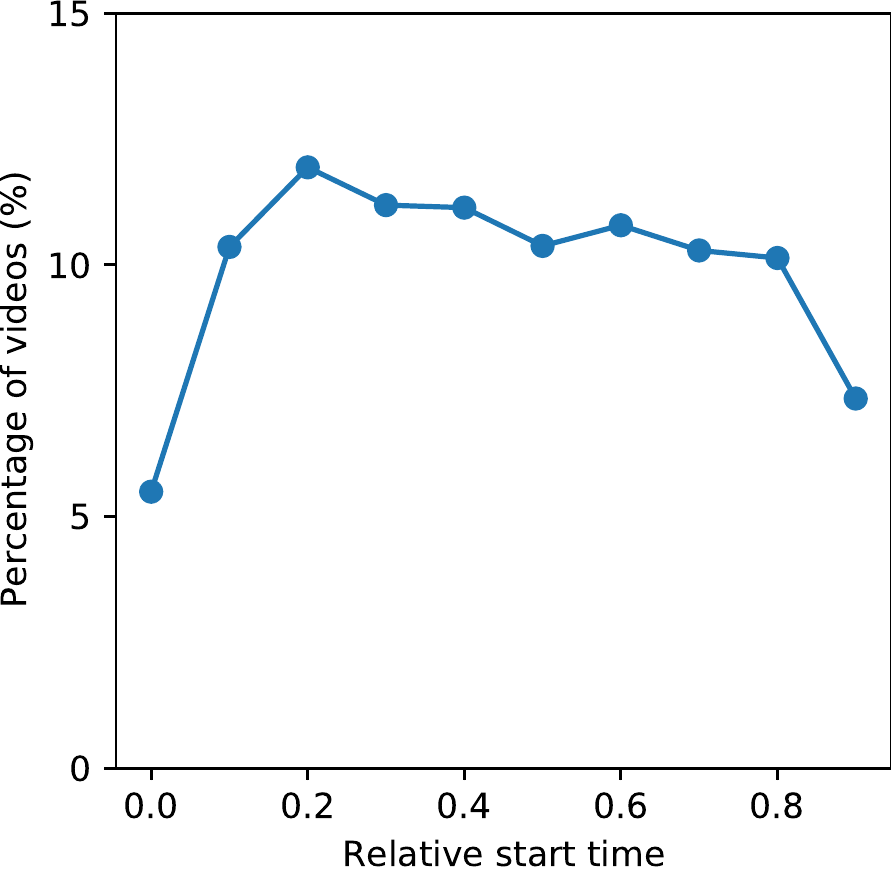}
\caption{Clip start time in the video}
\end{subfigure}%
\hfill
\begin{subfigure}{0.24\linewidth}
\vspace{-0.2cm}
\includegraphics[width=\linewidth]{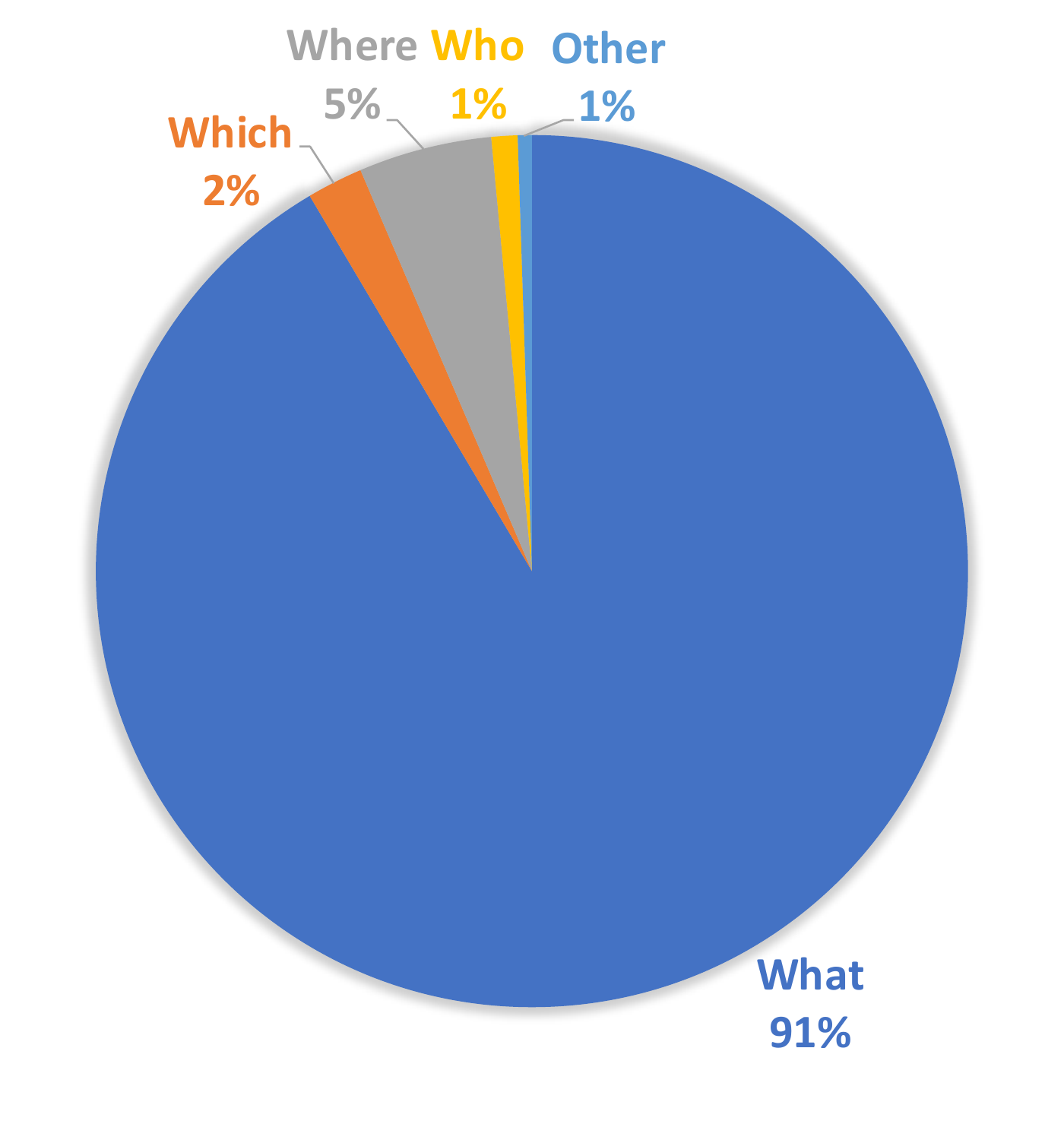}
\vspace{-0.7cm}
\caption{Question types}
\end{subfigure}
\vspace{-.2cm}
\caption{{\bf Statistics of the \smalldatasetname{} dataset.} (a)~Distribution of length of questions and answers. (b)~Distribution of video clip duration in seconds. (c)~Distribution of video clip relative start time in the original video. (d)~Distribution of question types.}
\label{fig:length}
\vspace{-.5cm}
\end{figure*}

\section{VideoQA model and training procedure}\label{sec:learning}
This section presents our VideoQA model (Section \ref{sec:model}) and describes the training procedure (Section \ref{sec:training}). Figure~\ref{fig:videoqamodel} gives an overview of the model.

\subsection{VideoQA model}\label{sec:model}

As illustrated in Figure \ref{fig:videoqamodel}, our VideoQA model is composed of two branches: \textit{(i)} a video-question module $f$ based on a transformer~\cite{vaswani2017attention} and a mapping from the CLS token with a linear function.
It takes a pair of video $v$ and question $q$ as input, models the multi-modal temporal interactions between $v$ and $q$ and then outputs an embedding vector $f(v,q) \in \nbR^{d}$. \textit{(ii)} The second branch is a text encoder $g$ that embeds an answer $a$ as $g(a) \in \nbR^{d}$. 
We will denote our model as \textit{VQA-T}, standing for VideoQA-Transformer. 
Note that using the joint (video, question) and answer embeddings allows us to deal with a large open vocabulary of answers present in our new  \datasetname{} dataset as the model can measure similarity between the input video-question embedding and the embedding of any answer. This is in contrast to using a classification answer module~\cite{huang2020location, jiang2020divide, jiang2020reasoning, le2020hierarchical, zhuang2020multichannel} that can choose only from a fixed predefined vocabulary of answers. Our embedding can be also easily finetuned on the different downstream VideoQA datasets, which may contain new answers that have not been seen at training. In contrast, the classification answer module has to be retrained when the vocabulary of answers changes. 
Next, we give details of the language and video representations. Further details about the model are provided in Appendix \ref{sec:mmt}. 

\vspace*{1mm}
\noindent \textbf{Word representation.}
The question and answer are separately tokenized with the WordPieces embedding~\cite{wu2016google} and fed to DistilBERT~\cite{sanh2019distilbert}. DistilBERT is a light version of BERT~\cite{bert18} pretrained in a self-supervised fashion on English Wikipedia and the Toronto Book Corpus \cite{zhu15aligning}.

\vspace*{1mm}
\noindent \textbf{Video representation.}
We use a frozen S3D~\cite{xie2018rethinking} pretrained on HowTo100M~\cite{miech19howto100m} using MIL-NCE~\cite{miech20endtoend}. This model is pretrained from scratch on HowTo100M only. 

\subsection{Training procedure}\label{sec:training}
This section describes the training of our VideoQA model on the \datasetname{} dataset and its finetuning on downstream VideoQA datasets.

\vspace*{1mm}
\noindent \textbf{Training on \datasetname{}.}
We wish to make a pair of video and question $(v,q)$ close to its correct answer $a$  measured by the dot product of their embeddings, $f(v,q)^\top g(a)$.
In contrast, the incorrect answers should be far, i.e., the dot product with their embeddings should be small. 
This can be done by maximizing the following contrastive objective:
\begin{equation}
\label{eq:objective}
\max_{f,g} \sum_{i=1}^n\log\left(\frac{e^{ f(v_i,q_i)^\top g(a_i)}}{e^{ f(v_i,q_i)^\top g(a_i)}+\sum\limits_{(v',q',a')\sim\mathcal{N}_i}e^{f(v',q')^\top g(a')}}\right),
\end{equation}
where $(v_i,q_i,a_i)$ represents a generated triplet (video clip, question, answer) from \datasetname{}.
Given a specific positive triplet $(v_i,q_i,a_i)$, we construct the set $\mathcal{N}_i$ of negative triplets by concatenating incorrect answers $a_j$ within the training batch to the video-question pair $(v_i, q_i)$ as: $(v_i,q_i,a_j)$ with $a_j \neq a_i$. 
In particular, if the same negative answer $a_j$ is present multiple times in a batch, we only count it once.
We found that sampling the same negative answer multiple times leads to worse results (see Section~\ref{sec:ablations}), which we believe is due to different distributions of answers in the pretraining and downstream datasets. Removing duplicate negatives helps to mitigate this difference.

\vspace*{1mm}
\noindent \textbf{Finetuning on downstream VideoQA datasets.}\label{sec:finetune}
We leverage the model pretrained on \datasetname{} and finetune it on a downstream VideoQA dataset that typically has a smaller vocabulary of answers $V$ (\eg $|V| \sim 4000$). 
To this end, we adapt the training objective in \eqref{eq:objective} by constructing the negative set $\mathcal{N}_i$ from {\em all} incorrect answers in $V$. Note that in such setting \eqref{eq:objective} becomes equivalent to optimizing the standard cross-entropy objective. In the specific case of multiple-choice VideoQA, the set of negatives $\mathcal{N}_i$ is the set of incorrect answers for each sample.

\vspace*{1mm}
\noindent \textbf{Masked Language Modeling (MLM).}\label{sec:mlm} 
In addition to the contrastive loss~(\ref{eq:objective}) we apply the masking loss~\cite{bert18} to question tokens during both pretraining and finetuning. We found this to have a positive regularization effect when finetuning the DistilBERT weights (see Section~\ref{sec:ablations}).

\section{\titlelowercase{i}VQA: a new VideoQA evaluation dataset}\label{sec:ivqa}
In this section we present our {\bf I}nstructional {\bf V}{\bf QA} dataset (iVQA). We start from a subset of HowTo100M videos and manually annotate video clips with questions and answers. 
We aim (i)~to provide a well-defined evaluation by including five correct answer annotations per question and (ii)~to avoid questions which can be answered without watching the video. 
The dataset is described below.

\vspace*{1mm}
\noindent \textbf{\smalldatasetname{} Data Collection.}\label{sec:collection}
iVQA videos are obtained by randomly sampling 7-30 sec.\ video clips from the HowTo100M dataset~\cite{miech19howto100m}. 
We avoid overlap between datasets and make sure iVQA and \datasetname{} have no videos in common.
Each clip is manually annotated with one question and 5 answers on Amazon Mechanical Turk.
We ask workers to annotate questions about objects and scenes in the video and remove videos that could not be annotated. 
The correctness of annotations is manually verified by the authors. Moreover, we manually reduce the language bias by excluding questions that could be answered without watching the video.
To increase diversity, each question is answered by 5 different workers. The answers are restricted to 4 words and are complemented by a confidence level. 
Questions that receive multiple answers with low confidence are removed.
We further describe our data collection interfaces in \cite{yang2021just} (Appendix C.1.).

\begin{figure}[t]
\centering
\begin{subfigure}{1.\linewidth}
\vspace{.2cm}
\begin{center}{\em Word cloud for questions\vspace{-.1cm}}\end{center}
\includegraphics[width=\linewidth]{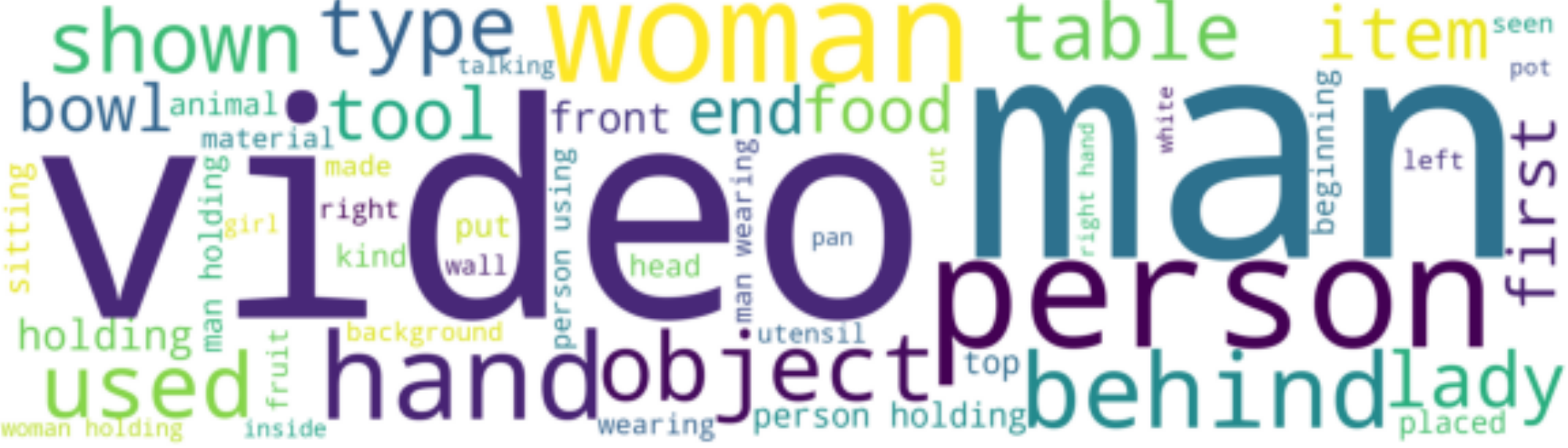}
\end{subfigure}
\begin{subfigure}{1.\linewidth}
\vspace{.2cm}
\begin{center}{\em Word cloud for answers\vspace{-.1cm}}\end{center}
\includegraphics[width=\linewidth]{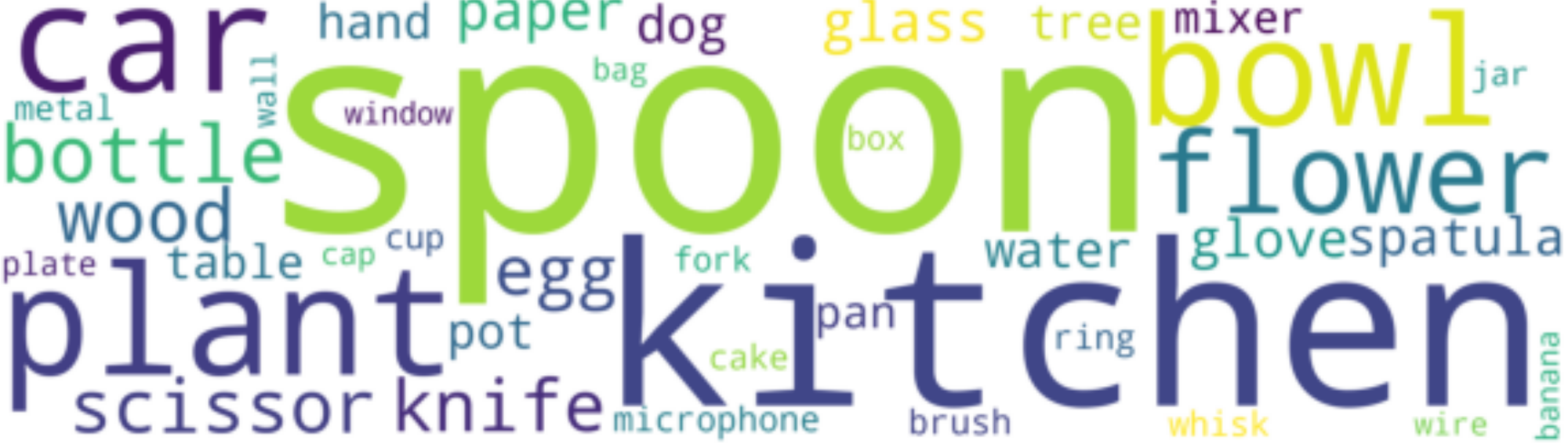}
\end{subfigure}
\caption{Word clouds for questions and answers in our \smalldatasetname{} dataset. 
The frequent  occurrence of location and time-specific words (\textit{behind}, \textit{front}, \textit{right}, \textit{left}, \textit{first}, \textit{end}, \textit{beginning}) indicates the presence of the spatial and temporal context within \smalldatasetname{} questions.
We can also observe the task-specific vocabulary in \smalldatasetname{} answers related to the domains of cooking, hand crafting and gardening.}
\vspace{-0.5cm}
\label{fig:words}
\end{figure}

\vspace*{1mm}
\noindent \textbf{Statistical analysis of \smalldatasetname{}.}\label{sec:analysis}
\smalldatasetname{} contains 10,000 video clips with one question and five corresponding answers per clip. We split the dataset into 60\%/20\%/20\% train/validation/test subsets. 
Figure \ref{fig:length} shows the distributions of question length, answer length, clip duration, clip relative start time in the original video and question types.
The average duration of video clips is 18.6 seconds.
Clip duration and start time distributions are almost uniform because we randomly sampled both the duration and the start time to obtain the clips, which results in a high video content diversity.
Most questions are about objects (\textit{What} questions make up 91\% of the data), while some are about places (\textit{Where} questions make up 5\% of the data) and people (\textit{Who} questions make up 1\% of the data).
On average, questions and answers contain 7.6 and 1.1 words, respectively.
Answers are in great majority one or two words, which is a result of our collection procedure.

The majority of questions have a consensus between at least 2 annotators, \ie at least 2 annotators providing the same answer. In detail, we observe that 27.0\% of questions lead to a perfect consensus among the five answer annotators, 48.4\% of questions lead to a consensus among at least four annotators, and 77.3\% lead to a consensus among at least three annotators.
All but six questions lead to a consensus between at least two annotators. 
Additionally, 27.5\% of questions have two different answers that had a consensus between at least two annotators. Similarly to~\cite{antol2015vqa}, this motivates us to define the following accuracy measure for a given answer $a$:
$acc(a)=\min(\frac{\#\textrm{ground truth answers =}\, a}{2},1).$
This metric assigns 100\% accuracy to answers confirmed by at least 2 annotators, 50\% accuracy to answers confirmed by only 1 annotator and 0\% otherwise. Note that this definition is specific to our set-up where we have \emph{multiple} ground truth answers per question.

Word clouds for questions and answers in the \smalldatasetname{} dataset in Figure~\ref{fig:words} demonstrate the relation of \smalldatasetname{} to the domains of cooking, hand crafting and gardening.
These word clouds also indicate that questions in \smalldatasetname{} often require spatial reasoning (\textit{behind}, \textit{front}, \textit{right}, \textit{left}) and temporal understanding (\textit{first}, \textit{end}, \textit{left}, \textit{beginning}) of the video. 
The most frequent answer (\textit{spoon}) in \smalldatasetname{} corresponds to 2\% of all answers in the dataset. 
In contrast, the most frequent answers in other existing VideoQA datasets account for more than 9\% of all answers in these datasets (we have verified this for MSRVTT-QA, MSVD-QA and ActivityNet-QA). 
As a consequence, the \textit{most frequent answer baseline} is significantly lower for our \smalldatasetname{} dataset compared to other VideoQA datasets. 
We further evaluate the language bias in \smalldatasetname{} in Section \ref{sec:bias}. 

\begin{table*}[t]
\vspace{-0pt}
\begin{center}
\resizebox{.9\linewidth}{!}{
\begin{tabular}{lc|ccccccccc}
Method & Pretraining Data & 
\multicolumn{2}{c}{\smalldatasetname{}} & \multicolumn{2}{c}{MSRVTT-QA} & 
\multicolumn{2}{c}{MSVD-QA} & 
\multicolumn{2}{c}{ActivityNet-QA} & 
How2QA
\\ 
& & Top-1 & Top-10 & Top-1 & Top-10 & Top-1 & Top-10 & Top-1 & Top-10 & Top-1 \\
\hline
Random & $\emptyset$
& 0.09 & 0.9 & 0.02 & 0.2 & 0.05 & 0.5 & 0.05 & 0.5 & 25.0 \\
\qat{} & \datasetname{}
& 4.4 & 23.2 & 2.5 & 6.5 & 4.8 & 15.0 & 11.6 & 45.8 & 38.4 \\
\vqat{} & HowTo100M &
1.9 & 11.9 & 0.3 & 3.4 & 1.4 & 10.4 & 0.3 & 1.9 & 46.2 \\
\vqat{} (Ours) & \datasetname{}
& \textbf{12.2} & \textbf{43.3} & \textbf{2.9} & \textbf{8.8} & \textbf{7.5} & \textbf{22.4} & \textbf{12.2} & \textbf{46.5} & \textbf{51.1} \\
\end{tabular}}
\vspace{-0.2cm}
\caption{\small Comparison with baselines for zero-shot VideoQA. Top-1 and top-10 (for open-ended datasets) accuracy are reported.}
\vspace{-.6cm}
\label{table:zeroshot}
\end{center}
\end{table*}

\section{Experiments}\label{sec:experiments}
This section demonstrates the benefits of training using our generated \datasetname{} dataset and compares our method to the state of the art.
We first outline the used datasets, baseline methods and implementation details in Section \ref{sec:protocol}.
We then present results for the novel zero-shot VideoQA task in Section~\ref{sec:zeroshot}. 
Next we present results for the novel VideoQA feature probe evaluation setting in Section~\ref{sec:probe}. 
The comparison to the state of the art in VideoQA and alternative training strategies is given in Section~\ref{sec:results}. Section~\ref{sec:rare} presents results for rare answers and split per question type.
Then we compare our VideoQA generation approach to previous methods in Section~\ref{sec:java}. We also apply our approach to another video-text datasets in Section~\ref{sec:webvid}. Finally, we show the importance of the visual modality in iVQA in Section~\ref{sec:bias} and present  ablation studies in Section~\ref{sec:ablations}.

\subsection{Evaluation Protocol}\label{sec:protocol}

\noindent \textbf{Datasets.}\label{sec:datasets}
We use three datasets for training and five datasets for evaluation as described below. 
We follow previous evaluation protocols for open-ended settings~\cite{le2020hierarchical} and use a fixed vocabulary of training answers.
Unless stated otherwise, we report top-1 test accuracy and use original splits for training, validation and test. 

For training we use our new \textbf{\datasetname{}} dataset introduced in Section~\ref{sec:\datasetname{}} with 90\% and 10\% videos in training and validation subsets. 
For comparison, we also train our model using a large-scale text-video dataset,~\textbf{HowTo100M}~\cite{miech19howto100m}, that contains videos with transcribed narrations but \emph{no video-question-answer} triplets.
Test and validation videos of downstream datasets are excluded from HowTo100M and \datasetname{}.
To evaluate the general applicability of our approach,  we generate another automatic VQA dataset based on \textbf{WebVid2M}~\cite{bain2021frozen}, which consists of 2.5M video-text pairs scraped from the web where video captions are obtained from readily-available alt-text descriptions, see Section~\ref{sec:webvid}.

We evaluate results on four open-ended VideoQA downstream datasets: \textbf{MSRVTT-QA}~\cite{xu2017video}, \textbf{MSVD-QA}~\cite{xu2017video}, \textbf{ActivityNet-QA}~\cite{yu2019activitynet} and our new \textbf{\smalldatasetname{}} dataset (see Section~\ref{sec:ivqa}).
We also evaluate on a multiple-choice VideoQA dataset \textbf{How2QA}~\cite{li2020hero} where each question is associated with one correct and three incorrect answers.
For MSRVTT-QA and MSVD-QA, we follow~\cite{le2020hierarchical} and use a vocabulary of the top $4000$ training answers for MSRVTT-QA, and all $1852$ training answers for MSVD-QA. For our \smalldatasetname{} dataset and ActivityNet-QA, we consider all answers that appear at least twice in the training set, resulting in $2348$ answers for \smalldatasetname{} and $1654$ answers for ActivityNet-QA.

\begin{figure}[!ht]
\centering
\includegraphics[width=1.\linewidth]{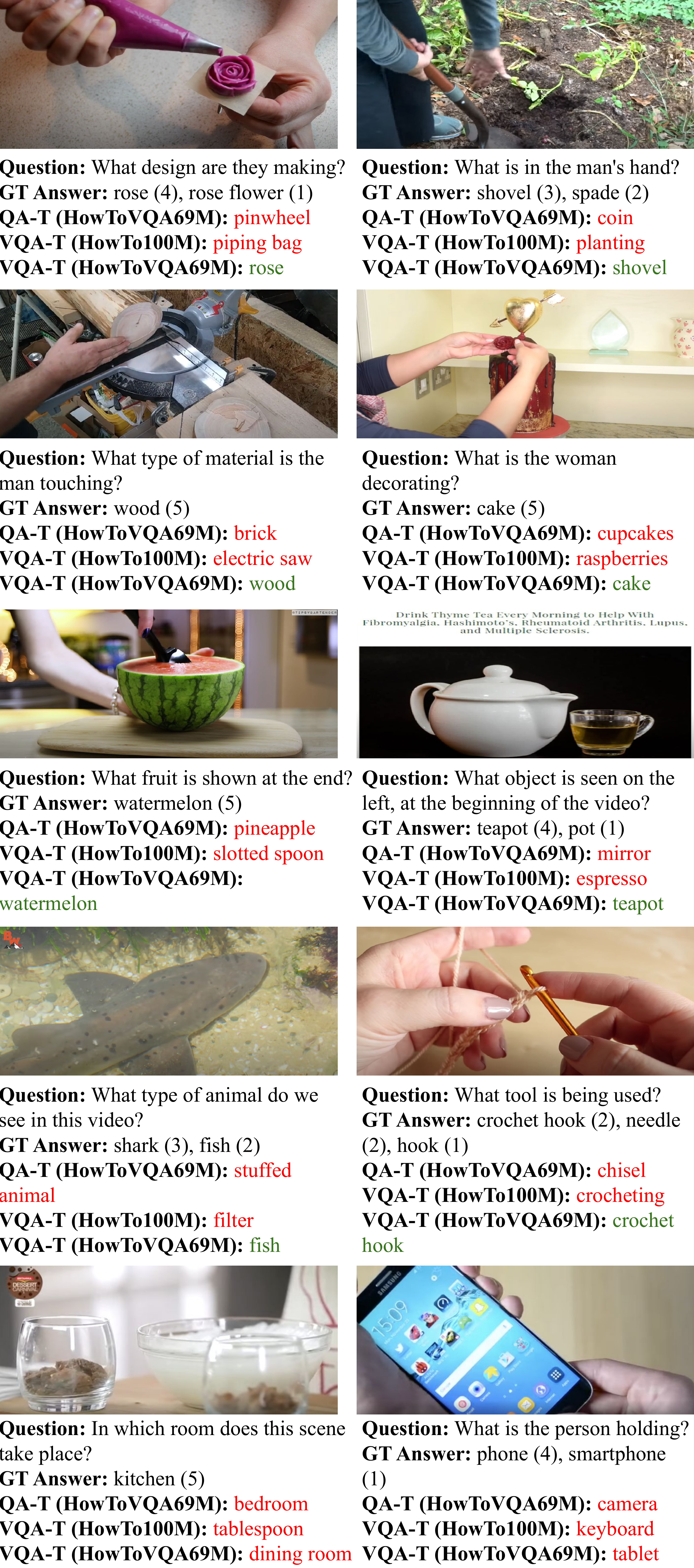}
\vspace{-0.5cm}
\caption{\small \textbf{Zero-shot VideoQA on \smalldatasetname{}}. The top 4 rows illustrate successful predictions of our model, while the bottom-most row displays failure cases. The values next to the ground truth (GT) answers indicate the number of annotators that gave the answer.
}
\label{fig:zeroshot}
\vspace{-0.5cm}
\end{figure}

\vspace*{1mm}
\noindent \textbf{Baselines.}\label{sec:baselines}
To evaluate the contribution of the visual modality, we compare our \textit{\vqat{}} model with its language-only variant \textit{\qat{}}.
\textit{\qat{}} does not use video input, i.e.~we set the input $v$ of the video-question transformer to zero during both training and testing (see Figure~\ref{fig:videoqamodel}).
To evaluate our generated dataset, we also compare \textit{\vqat{}} trained on~\datasetname{} and on HowTo100M.  
Since HowTo100M has no $(v,q,a)$ triplets, we only train the $f$ branch of \mbox{\textit{\vqat{}}} on HowTo100M using the standard masking and cross-modal matching losses \cite{chen2019uniter, li2020hero, lu2019vilbert, sun2019videobert, zhu2020actbert}. 
In the zero-shot setting we evaluate \textit{\vqat{}} trained on HowTo100M by computing $f(v,[q,a])$ for concatenated pairs of questions and answers $[q,a]$.
During finetuning we also initialize the $g$ branch of \textit{\vqat{}} with parameters of the text encoding obtained from $f$ (see further details in Appendix \ref{sec:mmt}). 

\vspace*{1mm}
\noindent \textbf{Implementation details.}\label{sec:details}
For the training on \datasetname{} we use the Adam optimizer and mini-batches with 4096 video clips sampled from 128 random videos. 
We use a cosine annealing learning rate schedule with initial value of $5 \times 10^{-5}$. 
The optimization over 10 epochs lasts 2 days on 8 Tesla V100 GPUs. 
Further details are included in Appendix \ref{sec:detailsbis}. 

\begin{table}[t]
\begin{center}
\setlength\tabcolsep{1pt}
\resizebox{\linewidth}{!}{	
\begin{tabular}{lc|ccccc}
Method & Pretraining data & \smalldatasetname{} & \makecell{ MSRVTT \\ QA} & \makecell{MSVD \\ QA} & \makecell{ActivityNet \\ QA} & How2QA \\
\hline
\vqat{} & $\varnothing$ & 3.8 & 23.2 & 21.8 & 22.9 & 55.3 \\
\qat{} & \datasetname{} & 11.4 & 27.0 & 29.5 & 27.6 & 64.7 \\
\vqat{} & HowTo100M & 13.8 & 27.0 & 32.9 & 24.7 & 63.9 \\ 
\vqat{} & \datasetname{} & \textbf{24.5} & \textbf{32.9} & \textbf{39.0} & \textbf{30.6} & \textbf{72.9} \\
\end{tabular}
}
\end{center}
\vspace{-0.4cm}
\caption{\small Probe evaluation of different pretraining strategies. In each case, only the last projection layers in the model were finetuned on the downstream VideoQA datasets. Top-1 accuracy is reported.}
\label{table:probe}
\end{table} 

\begin{table}[t]
\begin{center}
\setlength\tabcolsep{1.5pt}
\resizebox{\linewidth}{!}{	
\begin{tabular}{c|ccccc}
Pretraining data & \smalldatasetname{} & \makecell{ MSRVTT \\ QA} & \makecell{MSVD \\ QA} & \makecell{ActivityNet \\ QA} & How2QA \\
\hline
$\varnothing$ & 23.0 & 39.6 & 41.2 & 36.8 & 80.8 \\
HowTo100M & 28.1 & 40.4 & 43.5 & 38.1 & 81.9 \\ 
\datasetname{} & \textbf{35.4} & \textbf{41.5} & \textbf{46.3} & \textbf{38.9} & \textbf{84.4} \\
\end{tabular}
}
\end{center}
\vspace{-0.4cm}
\caption{\small Benefits of pretraining our \textit{VQA-T} model on our new HowToVQA69M dataset (last row) compared to no pretraining (first row) or pretraining on HowTo100M (second row). In each case our \textit{VQA-T} model was then finetuned on the downstream VideoQA datasets. Top-1 accuracy is reported.}
\vspace{-0.3cm}
\label{table:baselines}
\end{table} 

\subsection{Zero-shot VideoQA}\label{sec:zeroshot}

In this section, we address the {\em zero-shot VideoQA} task 
where we prohibit any manual supervision of visual data during training.
We explore this setup to evaluate the generalization of \textit{\vqat{}} trained on \datasetname{} to unseen downstream datasets.
For consistency, we use the vocabulary of answers from downstream datasets during testing (see Section~\ref{sec:datasets}). 

Zero-shot results are presented in Table \ref{table:zeroshot}. 
We first observe that the use of visual cues by \textit{\vqat{}} outperforms \mbox{\textit{\qat{}}} when both models are trained on \datasetname{}. 
This demonstrates the importance of the cross-modality in \datasetname{} despite the VideoQA annotation being exclusively generated from text-only methods.
Since \datasetname{} has been generated using no manual annotation of visual data, our approach is scalable and can lead to further improvements by increasing the dataset size, as we discuss in Section~\ref{sec:ablations}. 

\begin{table*}[t]
\centering
\setlength\tabcolsep{5pt}
\resizebox{1.\linewidth}{!}{
\begin{tabular}{cc|cccc|cccc|cccc|cccc}
Pretraining Data & Finetuning
& \multicolumn{4}{c}{iVQA} 
& \multicolumn{4}{c}{MSRVTT-QA} 
& \multicolumn{4}{c}{MSVD-QA} 
& \multicolumn{4}{c}{ActivityNet-QA} \\
& & Q1 & Q2 & Q3 & Q4 & Q1 & Q2 & Q3 & Q4 & Q1 & Q2 & Q3 & Q4 & Q1 & Q2 & Q3 & Q4 \\
\hline
$\varnothing$ & \cmark & 
38.4 & 16.7 & 5.9 & 2.6 &
\textbf{68.4} & 44.1 & 32.9 & 8.1 &
71.2 & 53.7 & 28.9 & 8.8 &
65.6 & 49.0 & 25.7 & 3.9 \\
HowTo100M & \cmark & 
46.7 & 22.0 & 8.6 & 3.6 &
65.2 & 46.4 & 34.9 & 10.6 &
\textbf{74.8} & 58.8 & 30.6 & 10.5 &
\textbf{67.5} & \textbf{53.3} & 25.9 & 4.1 \\
\datasetname{} & \xmark & 
9.0 & 8.0 & 9.5 & 7.7 &
0.2 & 6.4 & 2.4 & 3.0 &
9.3 & 9.0 & 6.9 & 4.8 &
36.3 & 5.7 & 3.7 & 1.5 \\
\datasetname{} & \cmark & 
\textbf{47.9} & \textbf{28.1} & \textbf{15.6} & \textbf{8.5} &
66.9 & \textbf{46.9} & \textbf{36.0} & \textbf{11.5} &
74.7 & \textbf{59.0} & \textbf{35.0} & \textbf{14.1} &
66.3 & 53.0 & \textbf{28.0} & \textbf{5.0} \\
\end{tabular}
}
\vspace{-0.3cm}
\caption{\small Results of our \textit{\vqat{}} model with different training strategies, on subsets of \smalldatasetname{}, MSRVTT-QA, MSVD-QA and ActivityNet-QA, corresponding to four quartiles with Q1 and Q4 corresponding to samples with the most frequent and the least frequent answers, respectively.}
\vspace{-0.2cm}
\label{table:rare}
\end{table*}

\begin{table*}[t]
\setlength\tabcolsep{2pt}
\begin{center}
\resizebox{.8\linewidth}{!}{	
\begin{tabular}{lc|cccccc|cccccc}
Pretraining Data & Finetuning & \multicolumn{6}{c}{MSRVTT-QA} & \multicolumn{6}{c}{MSVD-QA} 
\\ 
& & What & Who & Number & Color & When & Where
& What & Who & Number & Color & When & Where \\ \hline
$\varnothing$ & \cmark & 33.4 & 49.8 & 83.1 & 50.5 & 78.5 & 40.2 
& 31.5 & 54.9 & \textbf{82.7} & 50.0 & 74.1 & 46.4 \\
HowTo100M & \cmark & 34.3 & 50.2 & 82.7 & \textbf{51.8} & 80.0 & 41.5 
& 34.3 & 58.6 & 82.4 & \textbf{62.5} & \textbf{77.6} & \textbf{50.0} \\
\datasetname{} & \xmark & 1.8 & 0.7 & 66.3 & 0.6 & 0.6 & 4.5
& 7.8 & 1.7 & 74.3 & 18.8 & 3.5 & 0.0 \\
\datasetname{} & \cmark & \textbf{35.5} & \textbf{51.1} & \textbf{83.3} & 49.2 & \textbf{81.0} & \textbf{43.5} 
& \textbf{37.9} & \textbf{58.0} & 80.8 & \textbf{62.5} & \textbf{77.6} & 46.4 \\
\end{tabular}
}
\vspace{-0.3cm}
\caption{\small Effect of our pretraining per question type on MSRVTT-QA and MSVD-QA.}
\label{table:qtype}
\end{center}
\vspace{-0.4cm}
\end{table*}

\begin{table*}[t]
\setlength\tabcolsep{3pt}
\begin{center}
\resizebox{.8\linewidth}{!}{	
\begin{tabular}{lc|ccccccccc}
Pretraining Data & Finetuning & Motion & Spatial & Temporal & Yes-No & Color & Object & Location & Number & Other \\ \hline
$\varnothing$ & \cmark & 23.4 & 16.1 & 3.8 & 65.6 & 31.3 & 26.4 & 33.7 & 48.0 & 33.6 \\
HowTo100M & \cmark & 26.6 & \textbf{17.7} & 3.5 & \textbf{67.5} & 32.8 & 25.3 & 34.0 & \textbf{50.5} & 35.8 \\
\datasetname{} & \xmark & 2.3 & 1.1 & 0.3 & 36.3 & 11.3 & 4.1 & 6.5 & 0.2 & 4.7 \\
\datasetname{} & \cmark & \textbf{28.0} & 17.5 & \textbf{4.9} & 66.3 & \textbf{34.3} & \textbf{26.7} & \textbf{35.8} & 50.2 & \textbf{36.8} \\
\end{tabular}
}
\vspace{-0.3cm}
\caption{\small Effect of our pretraining per question type on ActivityNet-QA.}
\label{table:qtypeact}
\end{center}
\vspace{-0.6cm}
\end{table*}

\begin{table}[t]
\begin{center}
\setlength\tabcolsep{3pt}
\resizebox{\linewidth}{!}{	
\begin{tabular}{lc|ccc}
Method & Pretraining data & \makecell{MSRVTT \\ QA} & \makecell{MSVD \\ QA} \\
\hline
E-SA \cite{xu2017video} & & 29.3 & 27.6 \\ 
ST-TP \cite{jang2017tgif} & & 30.9 & 31.3 \\
AMU \cite{xu2017video} & & 32.5 & 32.0 \\
Co-mem \cite{gao2018motion} & & 32.0 & 31.7 \\ 
HME \cite{fan2019heterogeneous} & & 33.0 & 33.7 \\ 
LAGCN \cite{huang2020location} & & --- & 34.3 \\  
HGA \cite{jiang2020reasoning} & & 35.5 & 34.7 \\ 
QueST \cite{jiang2020divide} & & 34.6 & 36.1 \\ 
HCRN \cite{le2020hierarchical} & & 35.6 & 36.1 \\ 
MASN \cite{seo2021attend} & & 35.2 & 38.0 \\ 
Bridge to Answer \cite{park2021bridge} & & 36.9 & 37.2 \\ 
OCRL+LOGNet \cite{dang2021object} & & 36.0 & 38.2 \\
ClipBERT \cite{lei2021less} & \makecell{COCO \cite{chen2015microsoft}+ \\ Visual Genome \cite{visualgenome}}  & 37.4 & --- \\  
Jin \etal \cite{jin2021adaptive} & Conceptual Caption \cite{sharma2018conceptual} & 37.6 & 38.2 \\ 
SSML \cite{amrani2020noise} & HowTo100M & 35.1 & 35.1 \\
CoMVT \cite{seo2020look} & HowTo100M & 39.5 & 42.6 \\ 
SiaSamRea \cite{yu2021learning} & \makecell{COCO \cite{chen2015microsoft}+ \\ Visual Genome \cite{visualgenome}} & 41.6 & 45.5 \\ 
MERLOT \cite{zellers2021merlot} & YT-Temporal-180M & \textbf{43.1} & --- \\
\hline
\vqat{} & $\varnothing$ & 39.6 & 41.2 \\ 
\vqat{} & \datasetname{} & 41.5 & \textbf{46.3} \\
\vqat{} & \makecell{\datasetname{}+ \\ \webdataname{}} & 41.8 & \textbf{47.5} \\
\end{tabular}
}
\end{center}
\vspace{-0.5cm}
\caption{\small Comparison with state of the art on MSRVTT-QA and MSVD-QA (top-1 accuracy).}
\vspace{-0.4cm}
\label{table:sotamvtmvd}
\end{table}

\begin{table}[!htbp]
\setlength\tabcolsep{2.5pt}
\centering
\resizebox{.95\linewidth}{!}{	
\begin{tabular}{lc|cc}
& Pretraining data & \makecell{ActivityNet \\ QA} & How2QA \\
\hline
E-SA \cite{yu2019activitynet} & & 31.8 & --- \\
MAR-VQA \cite{zhuang2020multichannel} & & 34.6 & --- \\
HERO \cite{li2020hero} & \makecell{HowTo100M + \\ TV Dataset} & --- & 74.1 \\
CoMVT \cite{seo2020look} & HowTo100M & 38.8 & 82.3 \\
SiaSamRea \cite{yu2021learning} & \makecell{COCO \cite{chen2015microsoft}+ \\ Visual Genome \cite{visualgenome}} & 39.8 & 84.1 \\ 
MERLOT \cite{zellers2021merlot} & YT-Temporal-180M & \textbf{41.4} & --- \\

\hline
\vqat{} & $\varnothing$ & 36.8 & 80.8 \\
\vqat{} & \datasetname{} & 38.9 & \textbf{84.4} \\ 
\vqat{} & \makecell{\datasetname{}+ \\ \webdataname{}} & 39.0 & \textbf{85.3} \\
\end{tabular}
}
\vspace{-0.3cm}
\caption{\small Comparison with state of the art on ActivityNet-QA and the public val set of How2QA (top-1 accuracy).}
\vspace{-0.6cm}
\label{table:sotaacthow2qa}
\end{table}

Training on \datasetname{} significantly outperforms the training on HowTo100M and the random baseline. 
This confirms the advantage of our \datasetname{} dataset for the VideoQA task over other generic text-video datasets that do not contain video-question-answer triplets.
We emphasize that our training does not use any information about target VideoQA datasets.
Qualitative results for zero-shot VideoQA are presented for our approach and compared with baselines in Figure~\ref{fig:zeroshot}. 
We observe that \textit{\qat{}} (trained on \datasetname{}) provides plausible but video-unrelated answers to the questions. 
Moreover, \textit{\vqat{}} (trained on HowTo100M) is able to reply with answers related to the visual content, but doesn't take into account the question. 
Our \textit{\vqat{}} model trained on \datasetname{}, on the other hand, correctly understands questions and uses information in the video to provide correct answers, confirming results in Table~\ref{table:zeroshot}.
We also illustrate some failure cases in Figure~\ref{fig:zeroshot}, showing that our zero-shot \textit{\vqat{}} model can fail to understand fine variations in the video or the language, confusing a \textit{kitchen} with a \textit{dining room} or a \textit{phone} with a \textit{tablet}.

\subsection{VideoQA feature probe evaluation}\label{sec:probe}
In this section we further evaluate the generalization capabilities of the multi-modal representation learnt by our pretrained model.
To this end, we analyze the effect of \textit{\vqat{}} pretraining followed by the \emph{finetuning of the final projection layers} on the target datasets. More precisely, only the final MLP in the video-question module and the final linear layer in the answer module are finetuned. 
All other weights in the model (notably the video, question, answer representations and the multi-modal transformer) are kept frozen after the large-scale pre-training. 

Results for VideoQA feature probe evaluation are reported in Table~\ref{table:probe}. Similarly as for the zero-shot setting, we observe that the use of visual cues in \textit{\vqat{}} outperforms \mbox{\textit{\qat{}}} when both models are pretrained on \datasetname{}. 
Additionally, pretraining on \datasetname{} significantly outperforms the pretraining on HowTo100M and the probe baseline trained from scratch. 
Note that the probe baseline trained from scratch, despite having notably randomly initialized frozen multi-modal transformer weights, achieves reasonable absolute results as it can exploit dataset biases, which are further ablated in Section~\ref{sec:bias}.
Interestingly, we find that on the \smalldatasetname{} dataset, the probe evaluation of our model pretrained on \datasetname{} (24.5\% accuracy, first line in Table~\ref{table:probe}) outperforms the fully supervised model trained from scratch (23.0\% accuracy, first line in Table~\ref{table:baselines}).
These results further confirms the quality of our multi-modal representation  learnt from \datasetname{}.

\begin{figure*}[t]
\centering
\includegraphics[width=1.\linewidth]{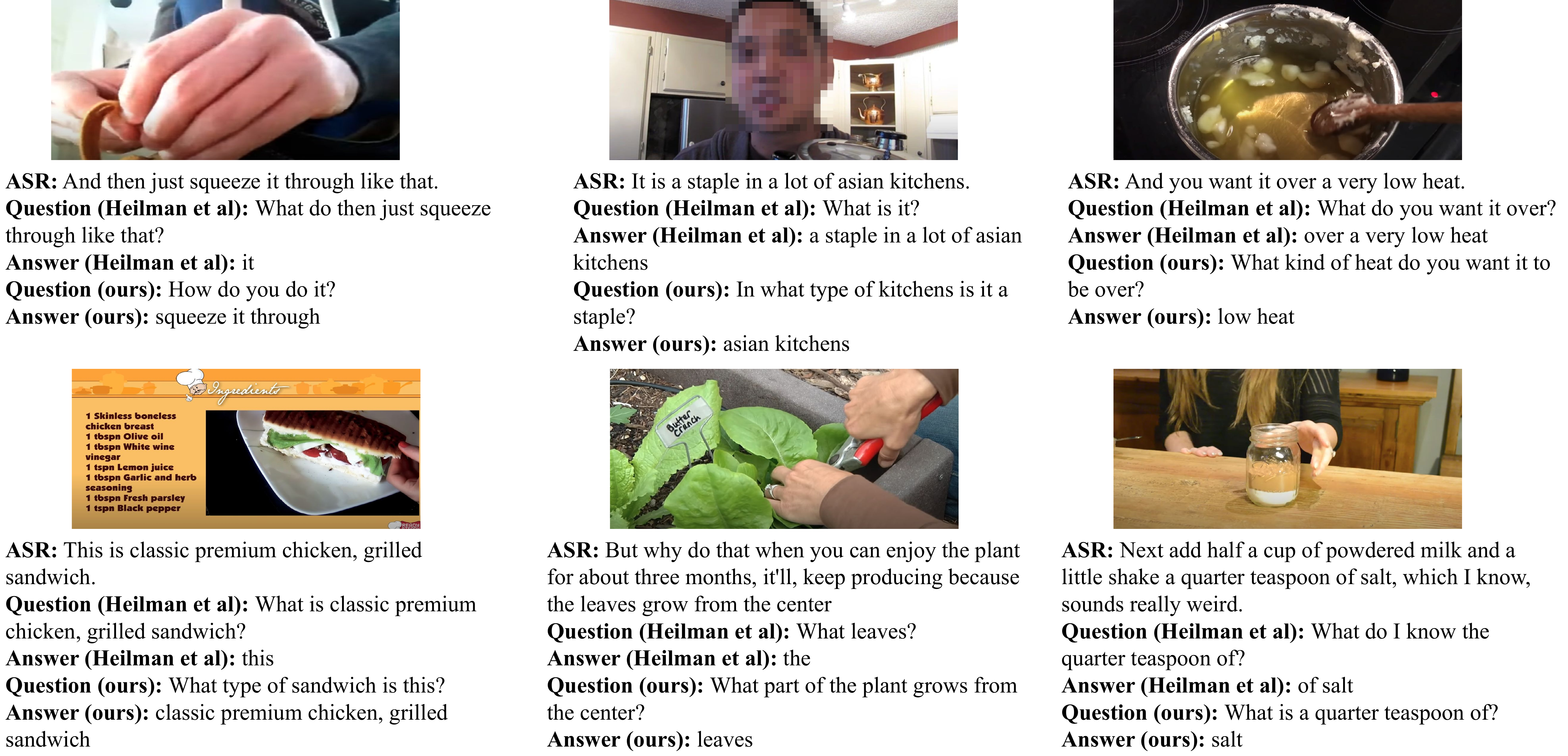}
\vspace{-0.7cm}
\caption{Qualitative examples of video-question-answer triplets generated with our trained language models compared to Heilman \etal~\cite{heilman2010good}, illustrating the higher quality and diversity of triplets obtained with our generation method.}
\vspace{-0.1cm}
\label{fig:java}
\end{figure*}

\begin{table*}[t]
\setlength\tabcolsep{4.5pt}
\begin{center}
\resizebox{0.95\linewidth}{!}{	
\begin{tabular}{c|cccc|cccc}
Pretraining Data & \multicolumn{4}{c}{Zero-shot} & \multicolumn{4}{c}{Finetune} \\
& \smalldatasetname{} & MSRVTT-QA & ActivityNet-QA & How2QA 
& \smalldatasetname{} & MSRVTT-QA & ActivityNet-QA & How2QA \\ 
\hline
$\varnothing$ & --- & --- & --- & ---
& 23.0 & 39.6 & 36.8 & 80.8 \\
MSRVTT-QA 
& 8.6 & --- & 1.7 & 42.5
& 25.2 & --- & 37.5 & 80.0 \\
ActivityNet-QA & 5.5 & 2.7 & --- & 40.8
& 24.0 & 39.9 & --- & 80.7 \\
\hline
\datasetname{} & \textbf{12.2} & \textbf{2.9} & \textbf{12.2} & \textbf{51.1}
& \textbf{35.4} & \textbf{41.5} & \textbf{38.9} & \textbf{84.4} \\
\end{tabular}
}
\vspace{-0.2cm}
\caption{\small Comparison of our training on \datasetname{} with cross-dataset transfer using the previously largest open-ended VideoQA dataset (MSRVTT-QA) and the largest manually annotated open-ended VideoQA dataset (ActivityNet-QA).}
\vspace{-0.6cm}
\label{table:transfer}
\end{center}
\end{table*}

\begin{table}[t]
\vspace{-0pt}
\setlength\tabcolsep{1.5pt}
\begin{center}
\resizebox{\linewidth}{!}{
\begin{tabular}{l|ccc|ccc}
\makecell{Generation \\ Method} & \multicolumn{3}{c}{Zero-shot} & \multicolumn{3}{c}{Finetune} \\
& \smalldatasetname{} & \makecell{ActivityNet \\ QA} & How2QA & \smalldatasetname{} & \makecell{ActivityNet \\ QA} & How2QA \\
\hline
\cite{heilman2010good} & 7.4 & 1.1 & 41.7 & 31.4 & 38.5 & 83.0 \\
Ours & \textbf{12.2} & \textbf{12.2} & \textbf{51.1} & \textbf{35.4} & \textbf{38.9} & \textbf{84.4} \\
\end{tabular}
}
\vspace{-0.2cm}
\caption{\small{Comparison of our question-answer generation approach with Heilman \etal~\cite{heilman2010good}, evaluated by downstream performance of the model trained on the generated VideoQA data.}}
\label{table:java}
\end{center}
\vspace{-0.5cm}
\end{table}

\subsection{Benefits of \datasetname{} pretraining}\label{sec:results}

This section evaluates the effect of \textit{\vqat{}} pretraining in combination with finetuning on target datasets.
As shown in Table~\ref{table:baselines}, pretraining on \datasetname{} provides consistent and significant improvements for all datasets when compared to pretraining on HowTo100M and no pretraining.
In particular, we observe the largest improvement for our new \smalldatasetname{} dataset which comes from the same domain as \datasetname{}. Hence, the automatic generation of training data for other domains using our method can lead to further improvements on other datasets.

We compare our pretrained model to the state-of-the-art in VideoQA in Tables~\ref{table:sotamvtmvd}-\ref{table:sotaacthow2qa}.
Notably, \textit{\vqat{}} pretrained on \datasetname{} outperforms previous methods using comparable pretraining data on all tested datasets. 
In particular, our method improves over CoMVT~\cite{seo2020look} that has been pretrained on HowTo100M.
We note that the recent SiaSamRea approach~\cite{yu2021learning} improves over our method on MSRVTT-QA (+0.1\%) and ActivityNet-QA (+0.9\%), but achieves lower results on MSVD-QA (-0.8\%) and How2QA (-0.3\%). However, SiaSamRea leverages manually annotated visual data for pretraining (COCO \cite{chen2015microsoft} and Visual Genome \cite{visualgenome}).
We also note that MERLOT~\cite{zellers2021merlot}  improves over our method on MSRVTT-QA and ActivityNet-QA, but uses the YT-Temporal-180M dataset for pretraining. This dataset includes HowTo100M but is significantly larger and more diverse (6 millions YouTube videos instead of 1 million).

\begin{figure*}[t]
\centering
\includegraphics[width=1.\linewidth]{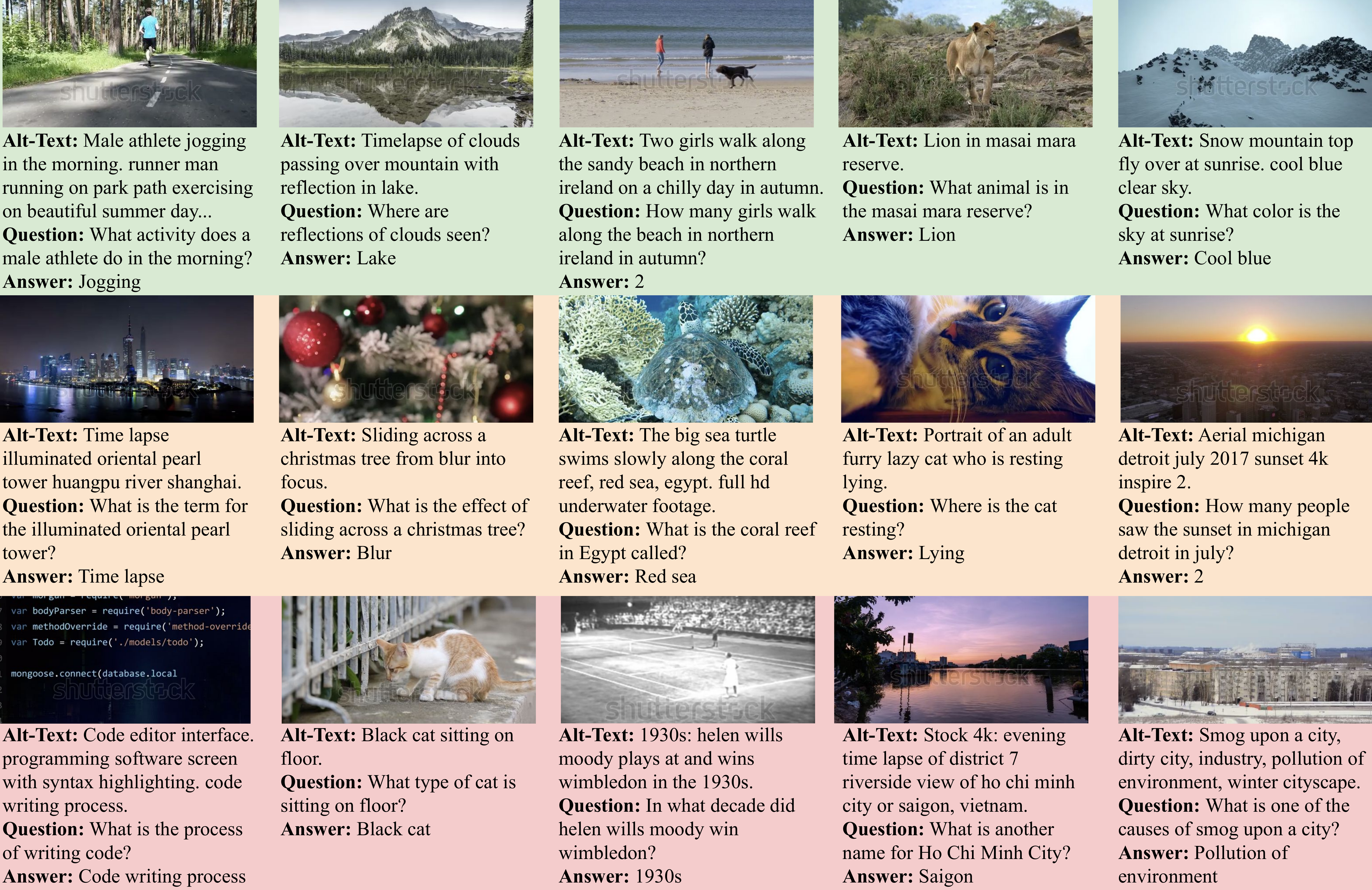}
\vspace{-0.7cm}
\caption{\small Examples of questions-answers generated from video alt-text pairs from the WebVid2M dataset~\cite{bain2021frozen}. {\color{green}The green color} (first row) indicates relevant examples, {\color{orange}the orange color} (second row) indicates a failure of the question-answer generation, and {\color{red}the red color} (third row) indicates that the generated question-answer is unrelated to the visual content.}
\vspace{-0.1cm}
\label{fig:webvid}
\end{figure*}

\begin{table*}[t]
\setlength\tabcolsep{3pt}
\begin{center}
\resizebox{1.\linewidth}{!}{	
\begin{tabular}{c|ccccc|ccccc}
Pretraining Data & \multicolumn{5}{c}{Zero-shot} & \multicolumn{5}{c}{Finetune} \\
& \smalldatasetname{} & MSRVTT-QA & MSVD-QA & ActivityNet-QA & How2QA 
& \smalldatasetname{} & MSRVTT-QA & MSVD-QA & ActivityNet-QA & How2QA \\ 
\hline
$\varnothing$ & --- & --- & --- & --- & ---
& 23.0 & 39.6 & 41.2 & 36.8 & 80.8 \\
\hline
\webdataname{}
& 7.3 & 5.3 & 12.3 & 6.2 & 49.8 &
28.1 & 41.2 & 45.4 & 38.1 & 82.4 \\
\datasetname{} & 12.2 & 2.9 & 7.5 & 12.2 & 51.1
& \textbf{35.4} & 41.5 & 46.3 & 38.9 & 84.4 \\
\hline
\makecell{\datasetname{} + \\ \webdataname{}} & \textbf{13.3} & \textbf{5.6} & \textbf{13.5} & \textbf{12.3} & \textbf{53.1}
& 35.2 & \textbf{41.8} & \textbf{47.5} & \textbf{39.0} & \textbf{85.3} \\
\end{tabular}
}
\vspace{-0.3cm}
\caption{\small Comparison of our VideoQA training datasets generated with different video-text data source, evaluated by downstream performance of the model pretrained on the generated data in zero-shot mode and after finetuning.}
\vspace{-0.5cm}
\label{table:webvid}
\end{center}
\end{table*}

\subsection{Analysis of rare answers and question types}\label{sec:rare}

\noindent \textbf{Results for rare answers.}
Training on downstream VideoQA datasets typically leads to particularly large improvements for questions with most frequent answers. 
As shown in Table \ref{table:rare}, our approach brings significant improvements both for common and rare answers compared to models trained from scratch or pretrained on HowTo100M.
We also find that our pretrained model, in the zero-shot setting, performs similarly across the different quartiles, with the exception of ActivityNet-QA, which includes in its most common answers \textit{yes}, \textit{no}.
Interestingly, for the most rare answers in \smalldatasetname{} (Q3 and Q4) our model without finetuning (zero-shot mode) outperforms finetuned models that have not been pretrained on \datasetname{}.
We conclude that VideoQA specific pretraining on additional large-scale, diverse data helps improve generalization of VideoQA models. 

Note that in order to have a consistent evaluation with other experiments, we keep the same train vocabulary at test time. 
This implies that a significant part of answers in the test set is considered wrong because the answer is not in the vocabulary. 
This represents 16\% of answers in \smalldatasetname{}, 3\% of answers in MSRVTT-QA, 6\% for MSVD-QA and 19\% for ActivityNet-QA. 
Note, however, that our joint embedding framework could allow for different vocabularies to be used at the training and test time. 

\vspace*{1mm}
\noindent \textbf{Results split per question type.} We also present results per question type for MSRVTT-QA, MSVD-QA and ActivityNet-QA in Tables \ref{table:qtype} and \ref{table:qtypeact}. Compared to the model trained from scratch or the model pretrained on HowTo100M, we observe consistent improvements by our model for most categories. 

\subsection{Comparison of VideoQA generation methods and VideoQA training datasets}\label{sec:java}

\noindent \textbf{Comparison of VideoQA generation methods.} We compare our question-answer generation approach to Heilman \etal~\cite{heilman2010good}, that was notably used in \cite{xu2017video, zeng2017leveraging, zhao2020open, zhao2017video,  zhao2018open} to generate VideoQA data from video descriptions.
We run the method of~\cite{heilman2010good} on sentences extracted from HowTo100M, apply our pretraining method on the generated data and show results in Table~\ref{table:java}. 
Note that we do not choose MSRVTT-QA and MSVD-QA as downstream datasets for this comparison because their evaluation sets were automatically generated using Heilman \etal~\cite{heilman2010good}. 
We find that our generation method leads to significantly better performance both in zero-shot and finetuning settings. 
We supplement this quantitative comparison with a qualitative comparison shown in Figure~\ref{fig:java}. 
We found that compared to~\cite{heilman2010good} our generation method provides higher quality as well as higher diversity of question-answer pairs when applied to the uncurated sentences extracted from speech in narrated videos.
This further demonstrates the benefit of our transformer-based question-answer generation approach compared to previous rule-based methods.

\noindent \textbf{Comparison of VideoQA training datasets.} We also evaluate the importance of our generated \datasetname{} dataset by comparing our results to cross-dataset transfer using existing VideoQA datasets. We define cross-dataset transfer as a procedure where we pretrain our VideoQA model on a VideoQA dataset and then finetune and test it on another VideoQA dataset. 
The training follows the procedure described for finetuning in Section \ref{sec:training}. 
We report results for cross-dataset transfer in Table \ref{table:transfer}. Note that we do not use MSVD-QA as downstream dataset as its test set has been automatically generated with the same method \cite{heilman2010good} as MSRVTT-QA.
As can be observed, our approach with pretraining on \datasetname{} significantly outperforms cross-dataset transfer models using the previously largest VideoQA dataset (MSRVTT-QA), or the largest manually annotated VideoQA dataset (ActivityNet-QA), both for the zero-shot and finetuning settings, on all four downstream datasets.
We emphasize that our dataset is generated relying on text-only annotations, while MSRVTT-QA was generated using manually annotated video descriptions and ActivityNet-QA was manually collected.
These results further demonstrate the benefit of our \datasetname{} dataset for training VideoQA models.

\begin{table}[t]
\begin{center}
\setlength\tabcolsep{4pt}
\resizebox{\linewidth}{!}{	
\begin{tabular}{l|ccccc}
Method & \smalldatasetname{} & \makecell{ MSRVTT \\ QA} & \makecell{MSVD \\ QA} & \makecell{ActivityNet \\ QA} & How2QA \\
\hline
\qat{}
& 14.1 & 32.8 & 32.6 & 30.4 & 76.6 \\
\vqat{} & 23.0 & 39.6 & 41.2 & 36.8 & 80.8 \\ 
\end{tabular}
}
\end{center}
\vspace{-0.4cm}
\caption{\small Comparison of \textit{\qat{}} and \textit{\vqat{}} models trained from scratch (without pretraining) on downstream datasets.}
\vspace{-0.3cm}
\label{table:bias}
\end{table} 

\subsection{Generalization to other video-text datasets}\label{sec:webvid} 
In this section, we show that our VideoQA generation approach can be generalized to other sources of non-manually annotated video-text paired data.
For this, we extend and apply our generation pipeline presented in Section~\ref{sec:qgen} to videos with alt-text description, \ie alt-text HTML attribute associated with videos, from the WebVid2M dataset~\cite{bain2021frozen}. 

\vspace*{1mm}
\noindent \textbf{\webdataname{} dataset.} We first explain how we adapt our generation pipeline detailed in Section~\ref{sec:qgen} to video alt-text pairs. 
As captions in WebVid2M are relatively short, we do not apply the punctuation model but directly apply the question-answer generation models on the captions. 
Captions in WebVid2M are also not temporally localized, so the generated question-answers are not temporally localized either.
They instead refer to the whole videos, which are typically short (4 seconds on average).
Applying our generation pipeline to WebVid2M~\cite{bain2021frozen}, we generate \webdataname{}, a dataset of 3,476,610 question-answers associated with 2,404,871 videos.
Examples of generated samples are illustrated in Figure \ref{fig:webvid}.
These examples show that despite a substantial visual-linguistic domain difference compared to HowTo100M, our approach is able to generate relevant VideoQA data. 
We believe that qualitatively, the generated QA data from \webdataname{} are of better quality than the generated QA data from \datasetname{} (see Section~\ref{sec:\datasetname{}}).
We argue that WebVid2M~\cite{bain2021frozen} has a better visual-linguistic correlation and a higher quality of text data compared to HowTo100M~\cite{miech19howto100m}, which facilitates the VideoQA generation.

\vspace*{1mm}
\noindent \textbf{Benefits of training on \webdataname{}.}
We next apply our pretraining method on the generated data and show results in Table~\ref{table:webvid}. We also explore combining both datasets with a simple curriculum learning strategy, where our model initially pretrained on \datasetname{} is further trained on \webdataname{}.
We find that training only on \webdataname{} gives competitive performance both in the zero-shot setting and the finetuning setting. Notably, it significantly improves over the variant trained from scratch in the finetuning setting. This shows that our approach can be generalized to other sources of video and text data. 
Additionally, we find that combining the two datasets for pretraining results in additional improvements both for zero-shot and finetuning.
Therefore the difference with previous methods is also increased (see Tables~\ref{table:sotamvtmvd}-\ref{table:sotaacthow2qa}).
Note that as \webdataname{} is significantly smaller than \datasetname{}, our training runs faster on this dataset (20 GPUH instead of 350 GPUH), which gives a practical advantage to \webdataname{}.
We have open-sourced \webdataname{} annotations to facilitate future research.

\subsection{Importance of the visual modality in \smalldatasetname{}}\label{sec:bias}

We show in Table~\ref{table:bias} that \textit{\qat{}} is a strong baseline compared to \textit{\vqat{}} on existing VideoQA datasets, when both are trained from scratch.
However, on \smalldatasetname{}, \textit{\vqat{}} improves even more over \mbox{\textit{\qat{}}} than with other datasets, as measured by absolute improvement in top-1 accuracy.
This suggests that the visual modality is more important in \smalldatasetname{} than in other VideoQA datasets. 

\subsection{Ablation studies}\label{sec:ablations}

\noindent \textbf{Pretraining losses.} As shown in Table~\ref{table:loss}, removing duplicate negative answers in our contrastive loss, as discussed in Section~\ref{sec:training}, is beneficial notably in the zero-shot setting.
Moreover, adding the MLM loss during pretraining improves the downstream results for both zero-shot and finetuning when used in combination with our contrastive learning strategy.
These results motivate our proposed pretraining approach.

\begin{table}[t]
\vspace{-0pt}
\setlength\tabcolsep{2pt}
\begin{center}
\resizebox{\linewidth}{!}{
\begin{tabular}{cc|cc|cc}
MLM & \makecell{ Sampling without \\ answer repetition} &
\multicolumn{2}{c}{Zero-shot} & \multicolumn{2}{c}{Finetune} \\
& & \smalldatasetname{} & MSVD-QA & 
\smalldatasetname{} & MSVD-QA \\
\hline
\xmark & \xmark & 11.1 & 6.1 & 34.7 & 45.6 \\
\xmark & \cmark & 12.1 & 7.0 & 34.3 & 45.0 \\
\cmark & \xmark & 10.9 & 6.4 & 34.3 & 45.1 \\
\cmark & \cmark & \textbf{12.2} & \textbf{7.5} & \textbf{35.4} & \textbf{46.3} \\
\end{tabular}
}
\vspace{-0.3cm}
\caption{\small{Effect of MLM loss and our negative sampling strategy on \datasetname{} training.}}
\vspace{-0.3cm}
\label{table:loss}
\end{center}
\end{table}

\vspace*{1mm}
\noindent \textbf{Importance of scale.}
Results of our method after pretraining on different fractions of \datasetname{} are shown in Table~\ref{table:scale}. 
We construct these subsets such that larger subsets include the smaller ones. 
These results suggest that the scale is an important factor and that we can expect further improvements with additional pretraining data, both in the zero-shot and finetuning settings.

\begin{table}[t]
\vspace{-0pt}
\setlength\tabcolsep{3pt}
\begin{center}
\resizebox{\linewidth}{!}{
\begin{tabular}{l|cc|cc}
Pretraining data size &
\multicolumn{2}{c}{Zero-shot} & \multicolumn{2}{c}{Finetune} \\
& \smalldatasetname{} & MSVD-QA & 
\smalldatasetname{} & MSVD-QA \\
\hline
0\% & --- & --- & 23.0 & 41.2 \\
1\% & 4.5 & 3.6 & 24.2 & 42.8 \\
10\% & 9.1 & 6.2 & 29.2 & 44.4 \\
20\% & 9.5 & 6.8 & 31.3 & 44.8 \\
50\% & 11.3 & 7.3 & 32.8 & 45.5 \\
100\% & \textbf{12.2} & \textbf{7.5} & \textbf{35.4} & \textbf{46.3}  \\
\end{tabular}
}
\vspace{-0.3cm}
\caption{\small{Effect of the training size of \datasetname{}.}}
\label{table:scale}
\end{center}
\vspace{-0.5cm}
\end{table}

\section{Conclusion}\label{sec:conclusion}
We propose a novel and scalable approach for training VideoQA models without manually annotated visual data.
We automatically generate \datasetname{} -- a large-scale VideoQA training dataset generated from narrated videos with readily-available speech transcripts, significantly exceeding existing datasets by size and diversity.
We demonstrate several benefits of pretraining on \datasetname{}. We are the first to demonstrate zero-shot VideoQA results while using no manually annotated images or videos for training. 
We also introduce the VideoQA feature probe evaluation setting and show strong generalization capabilities of the multi-modal representation learnt by our pretrained model.
Furthermore, finetuning our \datasetname{} pretrained model on downstream tasks achieves competitive performance on MSRVTT-QA, ActivityNet-QA, MSVD-QA and How2QA.
Moreover, we show that our approach generalizes to other sources of web videos by generating the \webdataname{} from video alt-text pairs and showing its benefits for VideoQA training.
We further validate our approach on our new manually-collected \smalldatasetname{} benchmark.



%

\ifCLASSOPTIONcompsoc
  \section*{Acknowledgments}
\else
  \section*{Acknowledgment}
\fi

This work was granted access to the HPC resources of IDRIS under the allocation 2020-101267 made by GENCI. The work was funded by a Google gift,  the French government under management of Agence Nationale de la Recherche as part of the "Investissements d'avenir" program, reference ANR-19-P3IA-0001 (PRAIRIE 3IA Institute), the Louis Vuitton ENS Chair on Artificial Intelligence and the European Regional Development Fund under project IMPACT (reg.\ no.\ CZ.02.1.01/0.0/0.0/15 003/0000468). 
We thank Pierre-Louis Guhur and Makarand Tapaswi for advice on using Amazon Mechanical Turk, Eloïse Berthier, Quentin Le Lidec and Elliot Chane-Sane for the manual evaluation of generated VideoQA data, and Ignacio Rocco for proofreading.



\ifCLASSOPTIONcaptionsoff
  \newpage
\fi



%



{\small
\bibliographystyle{ieee_fullname}
\bibliography{egbib}
}

\clearpage \newpage
\appendices
\section*{Appendix}
In this Appendix, we give additional architecture details for our VideoQA model in Section \ref{sec:mmt} and additional implementation details in Section \ref{sec:detailsbis}.

\section{VideoQA architecture}\label{sec:mmt}
\begin{figure}[t]
\centering
\includegraphics[width=1.\linewidth]{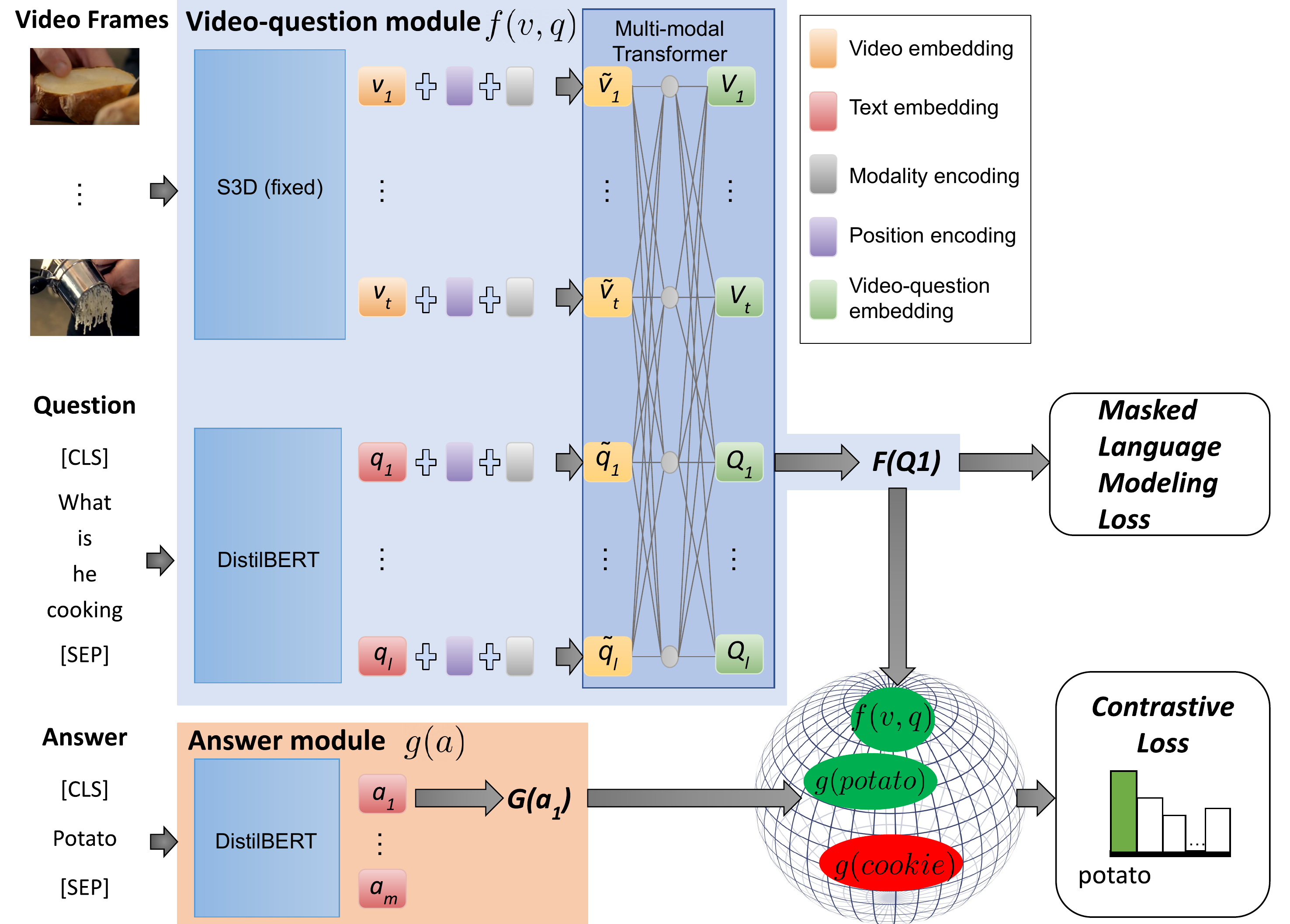}
\vspace{-0.5cm}
\caption{{\bf Overview of our VideoQA training architecture.} Our model is composed of a video-question module $f$ based on a multi-modal transformer (top) and an answer module $g$ based on DistilBERT \cite{sanh2019distilbert} encoder (bottom). For pretraining, we use a contrastive loss and a masked language modeling loss (right).}
\vspace{-0.5cm}
\label{fig:videoqamodelbis}
\end{figure}

Our architecture, detailed in Figure \ref{fig:videoqamodelbis}, has two main modules: (i) a video-question multi-modal transformer (top) and (ii) an answer transformer (bottom). Details are given next.

\vspace*{1mm}
\noindent \textbf{Video-question multi-modal transformer.} The input video representation, obtained from a fixed S3D model~\cite{xie2018rethinking}, is composed of $t$ features denoted  $v = [v_1, ..., v_t] \in \nbR^{d_v \times t}$ where $d_v$ is the dimension of the video features, and $t$ is the number of extracted features, one per second. 
The contextualized representation of the question, provided by the DistilBERT model~\cite{sanh2019distilbert}, is composed of $l$ token embeddings denoted as $q = [q_1, ..., q_{l}] \in \nbR^{d_q \times l}$ where $d_q$ is the dimension of the DistilBERT embedding and $l$ is the number of tokens in the question. 
The inputs to our video-question multi-modal transformer are then defined as a concatenation of question token embeddings and video features
\begin{equation}
\begin{split}
u(v, q) &= \left[\overset{\sim}{q}_1, ..., \overset{\sim}{q}_l, \overset{\sim}{v}_1, ..., \overset{\sim}{v}_t\right] \in \nbR^{d \times (l+t)},
\end{split}
\end{equation} 
with
\begin{equation}\overset{\sim}{q}_s = dp\left(\sigma\left(W_q q_s + b_q\right) + pos_s + mod_q\right),\end{equation} 
and 
\begin{equation}\overset{\sim}{v}_s = dp(\sigma(W_v v_s + b_v) + pos_s + mod_v),\end{equation}
where $W_q \in \nbR^{d_q \times d}$, $b_q \in \nbR^{d}$, $W_v \in \nbR^{d_v \times d}$, $b_v \in \nbR^{d}$ and learnable parameters, $mod_q \in \nbR^{d}$ and $mod_v \in \nbR^{d}$ are learnt modality encodings for video and question, respectively, and $[pos_1, ..., pos_{l+t}] \in \nbR^{d \times (l+t)}$ are fixed sinusoidal positional encodings. $\sigma$ is a Gaussian Error Linear Unit~\cite{hendrycks2016gaussian} followed by a Layer Normalization~\cite{ba2016layer} and $dp$ refers to Dropout~\cite{srivastava2014dropout}.

The multi-modal transformer is a transformer with $N$ layers, $h$ heads, dropout probability $p_d$, and hidden dimension $d_h$. The outputs of the multi-modal transformer $[Q_1, ... Q_l, V_1 ... V_t] \in \nbR^{d \times (l+t)}$ are contextualized representations over tokens in the question and temporal video representations. Finally, the fused video-question embedding $f(v,q)$ is obtained as 
\begin{equation}
F(Q_1) = W_{vq} dp(Q_1) + b_{vq},
\end{equation}
where $W_{vq} \in \nbR^{d \times d}$, $b_{vq} \in \nbR^{d}$ are learnable parameters and $Q_1$ is the multi-modal contextualized embedding of the [CLS] token in the question, as shown in Figure~\ref{fig:videoqamodel}.

\vspace*{1mm}
\noindent \textbf{Answer transformer.} The contextualized representation of the answer, provided by the DistilBERT model~\cite{sanh2019distilbert}, is composed of $m$ token embeddings denoted as $a = [a_1, ..., a_{m}] \in \nbR^{d_a \times m}$ where $d_a$ is the dimension of the DistilBERT embedding and $m$ is the number of tokens in the answer. 
Our answer embedding $g(a)$ is then obtained as
\begin{equation}
G(a_1) = W_a a_1 + b_a,
\end{equation}
where $W_{a} \in \nbR^{d_a \times d}$, $b_{a} \in \nbR^{d}$ are learnable parameters and $a_1$ is the contextualized embedding of the [CLS] token in the answer, as shown in Figure~\ref{fig:videoqamodelbis}.

\section{Additional experimental details}\label{sec:detailsbis}
Below we include additional details regarding the VideoQA generation, hyperparameter settings, training, masked language modeling and pretraining on HowTo100M.

\noindent \textbf{VideoQA generation.} The input sequence to the answer extractor and question generation transformers are truncated and padded up to a maximum of 32 tokens. 
The question decoding is done with beam search keeping track of the 4 most probable states at each level of the search tree. 
We have used the original captions (including stop words) from the HowTo100M dataset~\cite{miech19howto100m} and removed word repetitions from adjacent clips.

\noindent \textbf{VideoQA model.} We use the following hyperparameters: $l=20$, $t=20$, $m=10$, $d=512$, $d_h=2048$, $N=2$, $H=8$, $p_d=0.1$, $d_q=d_a=768$, $d_v=1024$.
The video features are sampled at equally spaced timestamps, and padded to length $t$. 
Sequences of question and answer tokens are truncated and padded to length $l$ and $m$, respectively. 
Attention is computed only on non-padded sequential video and question features. 

\noindent \textbf{Training.} 
For finetuning, we use a cosine annealing learning rate schedule with initial value of $1 \times 10^{-5}$. 
We use the Adam optimizer with batch size of 256 and training runs for 20 epochs. The final model is selected by the best performance on the validation set.

\noindent \textbf{Masked Language Modeling.} For the masked language modeling objective, a token is corrupted with a probability 15\%, and replaced 80\% of the time with [MASK], 10\% of the time with the same token and 10\% of the time with a randomly sampled token. 
To guess which token is masked, each sequential question output $Q_i$ of the multi-modal transformer is classified in a vocabulary of 30,522 tokens, and we use a cross-entropy loss.

\noindent \textbf{Pretraining on HowTo100M.} For video-text cross-modal matching, we sample one video negative and one text negative per (positive) video-text pair, and use a binary cross-entropy loss. 
The cross-modal matching module is used to perform zero-shot VideoQA for the variant \textit{VQA-T} trained on HowTo100M, by computing scores for $f(v,[q,a])$ for all possible answers $a$, for each video-question pair $(v,q)$.

\end{document}